\documentclass[11pt]{article}
\usepackage{xcolor}
\usepackage{geometry}                
\usepackage[parfill]{parskip}    

\usepackage[numbers]{natbib}
\usepackage{booktabs}
\usepackage{amssymb}
\usepackage{amsmath}
\usepackage{multicol}
\usepackage{graphicx}
\usepackage{array}
\usepackage{mathtools}
\usepackage{subfig}
\usepackage{multirow}
\usepackage{float}

\allowdisplaybreaks

\usepackage{listings}
\usepackage{color}

\usepackage{algorithm}
\usepackage{algcompatible}
\algnewcommand\INPUT{\item[\textbf{Input:}]}%
\algnewcommand\OUTPUT{\item[\textbf{Output:}]}%

\usepackage{titling}
\newcommand{\subtitle}[1]{%
	\posttitle{%
		\par\end{center}
	\begin{center}\large#1\end{center}
	\vskip0.5em}%
}

\def\exp{{\sf exp}}

\usepackage{xcolor}

\title{Morphology-Aware Ambiguity Learning for Wafer Defect Decision Support}

\author{
Seungjun Chu$^{1}$ and Seokhyun Chung$^{1,*}$\\[0.3em]
\small $^{1}$School of Industrial and Management Engineering,\\
\small Korea University, Seoul 02841, Republic of Korea\\[0.3em]
\small $^{*}$Corresponding author: \texttt{csh8901@korea.ac.kr}
}
\date{  }

\begin{document}

\maketitle

\begin{abstract}

Wafer map defect recognition is commonly formulated as a fixed-taxonomy classification problem that assigns each wafer to a single defect class. However, some wafers exhibit morphologies near class boundaries, for which forcing a single prediction may be less informative than providing plausible diagnostic alternatives. This paper proposes a morphology-aware ambiguity learning framework that supports three diagnostic actions: automatic single-class diagnosis, assisted diagnosis with two plausible defect classes, and full review. Using the radial, angular, and geometric characteristics of training wafer maps, the framework constructs a class-level ambiguity matrix representing defect-class pairs with similar morphology and plausible diagnostic alternatives. It guides the model to learn plausible alternative classes rather than treating all incorrect classes equally. During inference, the matrix determines whether an uncertain prediction can be represented by a meaningful two-class diagnostic set or should be escalated for full review. Experiments on WM-811K show that the proposed framework outperforms conventional approaches in defect recognition and diagnostic decision support, providing meaningful two-class alternatives while reserving full review for cases with unresolved ambiguity. Illustrative cost analyses further show the potential cost advantage of the proposed routing strategy. The diagnostic behavior of the framework remains consistent across different backbone architectures.

\end{abstract}



\section{Introduction}
\label{sec:introduction}

\begin{figure}[t]
    \centering
    
    \setlength{\tabcolsep}{4pt}
    \renewcommand{\arraystretch}{0.95}

    \begin{tabular}{cccc}
        \includegraphics[width=0.115\textwidth]{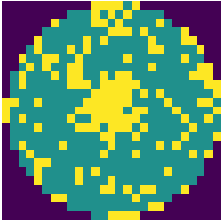} &
        \includegraphics[width=0.115\textwidth]{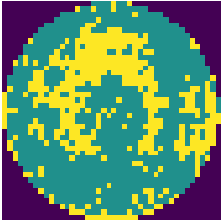} &
        \includegraphics[width=0.115\textwidth]{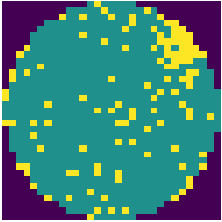} &
        \includegraphics[width=0.115\textwidth]{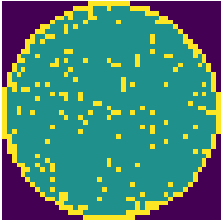}
        \\
        \small Center &
        \small Donut &
        \small Edge-Loc &
        \small Edge-Ring
    \end{tabular}

    \vspace{5pt}

    \begin{tabular}{ccccc}
        \includegraphics[width=0.115\textwidth]{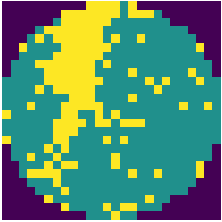} &
        \includegraphics[width=0.115\textwidth]{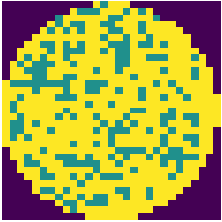} &
        \includegraphics[width=0.115\textwidth]{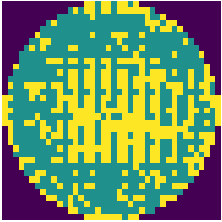} &
        \includegraphics[width=0.115\textwidth]{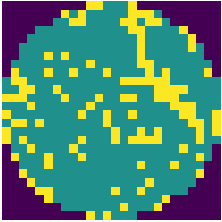} &
        \includegraphics[width=0.115\textwidth]{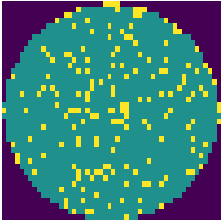}
        \\
        \small Loc &
        \small Near-full &
        \small Random &
        \small Scratch &
        \small Nonpattern
    \end{tabular}

    \vspace{5pt}
    \textbf{(a) Representative examples of the WM-811K classes}

    \vspace{10pt}

    \setlength{\tabcolsep}{2.5pt}
    \renewcommand{\arraystretch}{0.95}

    \begin{tabular}{cc@{\hspace{16pt}}cc}
        \multicolumn{2}{c}{\small Center $\leftrightarrow$ Donut} &
        \multicolumn{2}{c}{\small Loc $\leftrightarrow$ Scratch}
        \\[-1pt]

        \includegraphics[width=0.16\textwidth]{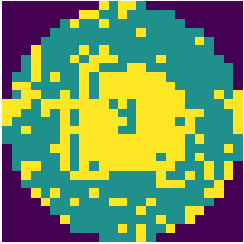} &
        \includegraphics[width=0.16\textwidth]{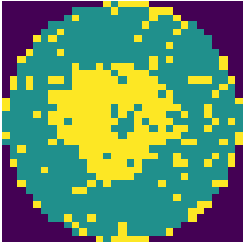} &
        \includegraphics[width=0.16\textwidth]{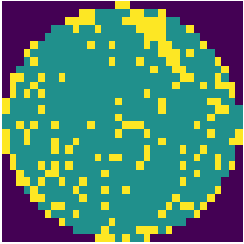} &
        \includegraphics[width=0.16\textwidth]{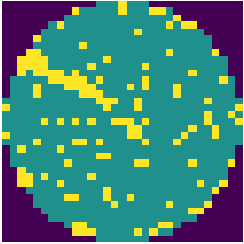}
        \\

        \small Center &
        \small Donut &
        \small Loc &
        \small Scratch
    \end{tabular}
    
    \vspace{5pt}
    \textbf{(b) Examples of morphology-level ambiguity}

    \caption{\small Representative wafer maps illustrating class taxonomy and boundary ambiguity. (a) Examples of the nine WM-811K classes; no class has a single canonical morphology. (b) Wafers from different annotated classes with visually similar defect morphologies.}
    \label{fig:taxonomy_ambiguity}
\end{figure}

Wafer maps have become an indispensable tool for process monitoring, root-cause analysis, and yield improvement in semiconductor manufacturing. They provide compact spatial summaries of die-level test outcomes, where spatial failure patterns often reveal underlying process abnormalities. These patterns enable experienced engineers to infer likely failure mechanisms and determine appropriate follow-up actions. However, as production volume and process complexity increase, manual inspection has become increasingly difficult to scale. This challenge has motivated the development of automatic wafer map analysis methods, including failure pattern classification and wafer similarity retrieval \cite{chen2000aneural,yuan2011detection,wu2015wafer}. Figure~\ref{fig:taxonomy_ambiguity}(a) shows nine representative defect classes from the widely used WM-811K benchmark \cite{wu2015wafer}.

Most learning-based methods formulate wafer map analysis as a single-label classification problem, where each wafer map is assigned to exactly one defect class. This formulation is simple, compatible with standard supervised learning objectives, and is therefore widely used in wafer map classification.
However, this assumption does not always hold in practice. Some wafer maps exhibit defect morphologies that share characteristics with multiple neighboring classes, making a unique class assignment difficult. Figure~\ref{fig:taxonomy_ambiguity}(b) illustrates this challenge using representative examples from the WM-811K benchmark. Although the dataset assigns each wafer map a single label for evaluation \cite{wu2015wafer}, wafer maps from different annotated classes can exhibit highly similar defect morphologies. For such boundary cases, a compact two-class diagnostic set may better represent the observed morphology than a single predicted label. Consequently, forcing a top-1 prediction may not always provide the most informative diagnosis.

Motivated by this observation, we reconsider the role of wafer defect classification in engineering diagnosis. Rather than forcing every wafer map into a single predefined defect class, our goal is to adapt the diagnostic action to the ambiguity of the observed defect pattern. 
Specifically, the proposed framework produces one of three diagnostic outcomes: (i) \textit{automatic diagnosis}, which returns a single label when the defect pattern is sufficiently characteristic of one class; (ii) \textit{assisted diagnosis}, which returns two plausible defect classes when the ambiguity can be explained by their morphological similarity; 
or (iii) \textit{full review}, which refers the wafer to a human engineer when the ambiguity remains unresolved.
This decision-support perspective allows the diagnostic action to reflect the nature of the ambiguity in each wafer's defect pattern.


A key distinction underlying this decision-support perspective is that low classifier confidence can arise for different reasons. In some cases, the classifier may simply have difficulty distinguishing among classes. In others, the observed defect morphology itself may plausibly correspond to more than one predefined class. We refer to the latter as \textit{boundary ambiguity}. In such cases, low confidence may reflect ambiguity in representing the observed morphology within the existing defect taxonomy, rather than only a limitation of the classifier.

Indeed, evidence of such boundary ambiguity is already present in the original WM-811K study. Wu \textit{et al.} \cite{wu2015wafer} reported difficulties in consistently annotating samples near class boundaries. For example, some wafer maps labeled as \texttt{Center} could also reasonably be interpreted as \texttt{Donut} or \texttt{Local}, while \texttt{Local} frequently overlaps with other defect patterns. These observations suggest that some apparent classification errors may arise from ambiguity between predefined defect classes rather than model error alone.


In this paper, we propose a morphology-aware ambiguity learning framework for wafer defect diagnosis. The key idea is to explicitly model which defect classes are plausible alternatives based on their morphological relationships, rather than treating all incorrect classes equally. To this end, we construct a class-level ambiguity matrix from the radial, angular, and geometric characteristics of training wafer maps. Each entry represents the extent to which two defect classes exhibit similar morphology, thereby they can form a plausible diagnostic alternatives.



The ambiguity matrix is used in both training and inference. During training, it encourages the classifier to learn morphologically plausible alternatives to the annotated class. During inference, the same matrix determines whether the two leading predictions form a plausible diagnostic pair. The resulting framework routes each wafer to one of three actions: automatic single-class diagnosis for sufficiently clear cases, assisted diagnosis for morphologically interpretable ambiguity, and escalation for full review when the ambiguity remains unresolved. In this way, morphology provides a common basis for both learning class relationships and determining the appropriate diagnostic action.

Our experiments demonstrate that the proposed framework improves both defect recognition and diagnostic decision support. Morphology-aware training improves defect classification compared with conventional and morphology-blind alternatives, while morphology-aware routing provides more meaningful two-class diagnostic sets than confidence-based routing alone. The framework also shows potential operational benefits under scenario-based cost analysis and maintains consistent diagnostic behavior across different backbone architectures.



The contributions of this paper are summarized as follows.
\begin{itemize}
    \item We identify morphology-level ambiguity as a distinct bottleneck in single-label wafer map classification. This bottleneck arises because some wafer maps naturally lie between predefined defect classes and cannot always be adequately represented by a single-label diagnosis.
    \item We propose a morphology-derived ambiguity matrix $\mathbf{A}^{\mathrm{morph}}$ that identifies which pairs of defect classes are morphologically plausible, using radial, angular, and geometric descriptors of wafer defect patterns. 
    \item We develop an ambiguity-aware training objective and an inference rule that suggests one of three operational actions: automatic diagnosis (single-class), assisted diagnosis (two-class), and escalation to full review.
    \item We validate the proposed framework on the WM-811K benchmark through comprehensive comparisons with representative classification and uncertainty-aware baselines. The results demonstrate that our framework improves diagnostic decision support by distinguishing morphologically plausible two-class cases from ambiguous cases that require full review.
\end{itemize}


The remainder of the paper is organized as follows. Section~\ref{sec:related_work} reviews related literature on wafer map diagnosis and ambiguity-aware learning. Section~\ref{sec:methodology} presents the proposed morphology-aware ambiguity learning framework. Section~\ref{sec:experiments} describes the experimental setup and evaluates the proposed framework. Finally, Section~\ref{sec:conclusion} concludes the paper with limitations and directions for future research.

\section{Related Work}
\label{sec:related_work}

\subsection{Wafer Map Defect Recognition}
\label{subsec:rw_wm_recognition}

Wafer map defect recognition has been studied as a scalable alternative to manual inspection in semiconductor manufacturing. Early studies developed neural-network-based recognition systems \cite{chen2000aneural}, model-based approaches for detecting spatial defect patterns \cite{yuan2011detection}, online detection and classification methods \cite{chien2013online}, and feature-based systems for large-scale wafer map analysis \cite{wu2015wafer}. In particular, Wu \textit{et al.} \cite{wu2015wafer} introduced the WM-811K dataset and developed rotation- and scale-invariant features for failure pattern recognition and similarity ranking. This work helped establish wafer map analysis as both a classification and retrieval problem for manufacturing decision support.


Subsequent research has substantially improved wafer map recognition through advances in representation learning. Convolutional neural networks (CNN)-based methods have been developed for defect classification and retrieval \cite{nakazawa2018wafer, yu2019wafer}, while later studies introduced rotation-invariant architectures \cite{kang2020rotation} and self-supervised representation learning \cite{kahng2021self, kwak2023swaco}. Collectively, these studies have improved the ability of learning-based models to extract informative representations of spatial defect patterns.

Another major challenge is the severe class imbalance of wafer map datasets such as WM-811K, where several defect types have relatively few labeled examples. Prior studies have addressed this problem through imbalance-aware learning and data augmentation, including generative approaches \cite{wang2019AdaBalGAN}, deep learning under imbalanced data \cite{saqlain2020deep}, and augmentation-based classification methods \cite{tsai2020light}. These approaches primarily aim to improve recognition performance for underrepresented defect classes.

Despite these advances, most existing studies formulate wafer map recognition as a fixed-taxonomy classification problem in which each wafer is assigned a single defect class. Accordingly, the primary objective is to improve the correctness of this single-class prediction through better representation learning, classifier architectures, or training strategies. This formulation does not explicitly account for cases in which the observed morphology lies near the boundary between defect categories.

\subsection{Uncertain Labels and Boundary Cases in Wafer Map Analysis}
\label{subsec:rw_ambiguity}

Evidence of class-boundary ambiguity can already be found in the construction and analysis of existing wafer map datasets. Wu \textit{et al.} \cite{wu2015wafer} report that wafer maps near classification boundaries can be difficult to assign to a single defect class. For example, some wafers labeled as \texttt{Center} could also reasonably be interpreted as \texttt{Donut} or \texttt{Local}, and \texttt{Local} was frequently confused with other defect types. Importantly, some of these alternative predictions were considered acceptable because the corresponding wafer maps appeared to lie near the boundary between two defect types. These observations suggest that some disagreements in classification may reflect genuine morphological ambiguity rather than simply classification error.


Related studies have considered several other forms of uncertainty in wafer map recognition. Park \textit{et al.} \cite{park2021discriminative} addressed uncertain annotations through defect label reconstruction, where the objective is to correct or refine unreliable labels. Jang \textit{et al.} \cite{jang2020support} considered open-set recognition, where a wafer may belong outside the predefined label set. Lee and Kim \cite{lee2020semi} studied mixed-type defects, where multiple defect patterns coexist within a wafer and a multi-label representation is thus appropriate. Kim \textit{et al.} \cite{kim2020uncertain} considered uncertainty in spatial features derived from
multiple wafer maps, including defect shape and location, while Liao \textit{et al.}
\cite{liao2026evidential} more recently addressed data uncertainty arising from mixed defect
morphologies and random noise through feature-level evidential modeling. While these problems all challenge conventional closed-set single-label classification, they represent different sources of uncertainty.  

Instead, our work focuses on boundary ambiguity within a fixed defect taxonomy. The candidate classes remain known. The wafer need not contain multiple distinct defect patterns. Rather, the ambiguity arises when the observed morphology can plausibly lie between two existing defect categories. In such cases, the relevant diagnostic question is whether a pair of morphologically plausible classes provides sufficient information without requiring full review.

\subsection{Soft Supervision and Morphology-Aware Class Relations}
\label{subsec:rw_soft_morphology}

Soft supervision provides a natural way to relax hard one-hot classification targets when alternative labels may also carry useful information. For example, label smoothing (LS) distributes a small amount of probability mass to off-target classes to reduce overconfidence \cite{szegedy2016rethinking}, while label distribution learning represents an instance by a distribution over labels to capture different degrees of label relevance \cite{geng2016label,gao2017deep}. These approaches demonstrate that supervision need not treat the annotated class as the only informative target.

However, applying soft supervision to wafer map diagnosis raises an important question: \emph{which alternative classes should receive probability mass?} Standard LS distributes probability mass without considering relationships among classes, while label distribution learning generally requires information about how strongly alternative labels describe each instance. Such information is not available in typical existing wafer datasets (e.g., WM-811K), which provide a single expert annotation for each labeled wafer \cite{wu2015wafer}. Generic soft supervision therefore does not by itself identify which alternative defect classes are diagnostically plausible.

Wafer morphology provides a natural source of structure for identifying such class relations. Prior wafer-map studies have characterized defect patterns using spatial and geometric attributes such as radial location, regional defect density, coverage, eccentricity, connectivity, and linearity \cite{wu2015wafer,jin2019novel}. These descriptors capture how defect classes differ in their spatial organization and can therefore provide information about which classes exhibit similar morphology.

Motivated by these two lines of research, our framework uses morphological structure to guide soft supervision. Rather than distributing off-target probability mass uniformly, we instead construct a class-level ambiguity matrix from classifier-independent morphological descriptors to identify structurally plausible class relations. This matrix is then used to shape the training targets, such that off-target probability mass reflects plausible morphological alternatives rather than arbitrary incorrect classes.

\subsection{Selective Classification and Set-Valued Prediction}
\label{subsec:rw_selective}

Selective classification allows a prediction system to abstain when its prediction is unreliable, trading prediction coverage for improved reliability \cite{chow1970on,geifman2017selective,geifman2019SelectiveNet}. This idea has also been applied to wafer map analysis. Alawieh \textit{et al.} \cite{alawieh2020wafer} applied deep selective learning to wafer defect classification, while Yoon and Kang \cite{yoon2022semi} used predictive uncertainty to determine whether a wafer should be automatically classified or referred to a process engineer. These approaches demonstrate the value of selective prediction for wafer diagnosis, but primarily consider a two-way decision between automatic classification and human review.

Set-valued prediction provides another way to handle uncertain predictions by returning multiple candidate classes. In particular, conformal prediction constructs prediction sets with statistical coverage guarantees under appropriate assumptions \cite{vovk2005Algorithmic,shafer2008atutorial}. While such sets provide a principled way to retain multiple possible labels, they do not necessarily indicate whether the included classes form a meaningful diagnostic pair. For wafer diagnosis, this distinction is important because uncertainty between two morphologically related defect classes may be more informative than uncertainty spread across unrelated classes.

Our framework bridges these two perspectives by introducing an intermediate diagnostic action between automatic classification and full review. When a single-class prediction is sufficiently reliable, the wafer is automatically diagnosed; when the uncertainty is concentrated between two morphologically plausible classes, the framework returns the pair as diagnostic alternatives; otherwise, the wafer is escalated for human review. Thus, rather than treating all uncertain predictions equally, the framework uses morphological structure to determine when uncertainty itself can provide useful diagnostic information.

\section{Proposed Approach}
\label{sec:methodology}
 
\subsection{Problem Formulation}
\label{subsec:problem}
 
Let $\mathcal{X}=\{0,1,2\}^{H\times W}$ denote the space of wafer maps, and let $x\in\mathcal{X}$ denote an individual wafer map, where each pixel value represents an outside-wafer location, an intact chip, or a defective chip, respectively. Each wafer map is assigned a class label $y \in \mathcal{Y}=\{0,\dots,K-1\}$ with $K=9$. Label $0$ corresponds to the \texttt{Nonpattern} class, and labels $1,\dots,8$ correspond to the eight named defect patterns: \texttt{Center}, \texttt{Donut}, \texttt{Edge-Loc}, \texttt{Edge-Ring}, \texttt{Loc}, \texttt{Near-full}, \texttt{Random}, and \texttt{Scratch}. We denote the eight defect classes by $\mathcal{Y}_d=\{1,\dots,8\}$. Throughout the paper we use the fixed class indexing summarized in Table~\ref{tab:class_index}.

\begin{table}[H]
\centering
\caption{Fixed class indexing used throughout the paper. Labels $1$--$8$ constitute the defect block $\mathcal{Y}_d$.}
\label{tab:class_index}
\begin{tabular}{cll}
\hline
$y$ & Class name & Diagnostic role \\
\hline
0 & \texttt{Nonpattern} & No identifiable failure pattern \\
1 & \texttt{Center}     & Defect concentrated at wafer center \\
2 & \texttt{Donut}      & Ring-shaped defect around the center \\
3 & \texttt{Edge-Loc}   & Localized cluster on the wafer edge \\
4 & \texttt{Edge-Ring}  & Full peripheral ring \\
5 & \texttt{Loc}        & Localized cluster away from the edge \\
6 & \texttt{Near-full}  & Defect covering most of the wafer \\
7 & \texttt{Random}     & Spatially scattered defects \\
8 & \texttt{Scratch}    & Elongated line-shaped defect \\
\hline
\end{tabular}
\end{table}
 
A standard classifier $f_\theta:\mathcal{X}\to\Delta^{K-1}$ maps an input wafer map $x$ to a class-probability vector $p_\theta(\cdot\mid x)$ and is trained using cross-entropy (CE),
\begin{equation*}
\mathcal{L}_{\mathrm{CE}}(\theta)
=
-\mathbb{E}_{(x,y)}
\left[
\log p_\theta(y\mid x)
\right].
\end{equation*}
This formulation treats the dataset label as the single target used for both training and evaluation. For many wafers, this assumption is appropriate. For wafers near the boundary between defect classes, however, a single top-1 prediction may not provide the most useful diagnosis. When two defect classes exhibit similar visual and morphological characteristics, reporting both candidates can provide more useful diagnostic information than committing to a single class, which may conceal the underlying ambiguity.

To this end, we formulate wafer defect diagnosis as a three-level decision process. Depending on the predicted ambiguity, the model selects one of three diagnostic outcomes:

\begin{itemize}
    \item [(i)] automatic diagnosis with a single defect label $\{\hat y\}$,
    \item [(ii)] assisted diagnosis with a two-class diagnostic set $\{ \hat y_1, \hat y_2 \}$, or
    \item [(iii)] escalation for full review.
\end{itemize}

The first action returns a single class when the observed morphology provides clear evidence for one diagnosis. The second action returns a two-class diagnostic set when the defect morphology is consistent with a known boundary between two classes. The third action escalates the wafer for full review when the observed ambiguity cannot be adequately explained by the predefined morphological relationships. The goal is therefore to design a framework to build a classifier that not only predicts defect classes, but also determines the appropriate level of diagnostic information to provide for each wafer.  

\subsection{The Framework}
\label{subsec:overview}

\begin{figure}[t]
\centering
\includegraphics[width=1.0\linewidth]{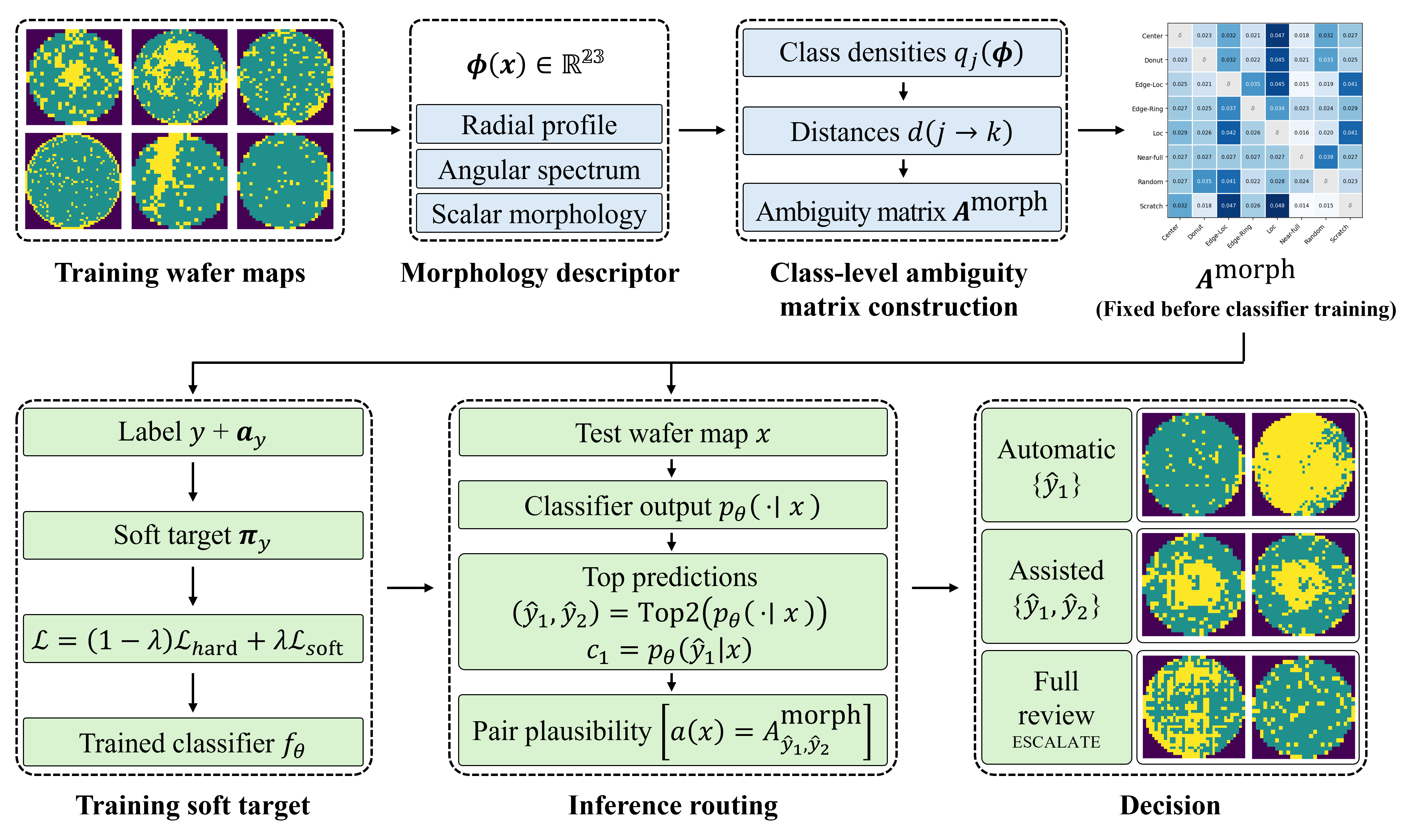}
    \caption{Overview of the proposed framework. The ambiguity matrix $\mathbf{A}^{\mathrm{morph}}$ derived from wafer morphology plays a key role in both training and inference. During training, it defines morphology-aware soft targets. During inference, it determines whether the top-2 predictions form a plausible two-class diagnostic set.}
    \label{fig:method_overview}
\end{figure}

We now discuss our proposed framework. The central idea is to distinguish morphologically explainable ambiguity from arbitrary classification uncertainty. Figure~\ref{fig:method_overview} summarizes the proposed framework. 

Our framework first identifies which inter-class confusions are supported by the underlying defect morphology. To this end, we extract deterministic morphological descriptors $\boldsymbol{\phi}(x)$ that capture the radial, angular, and geometric characteristics of each wafer map (Section \ref{subsec:morphology_descriptors}). These descriptors are then used to construct a morphology-derived ambiguity matrix $\mathbf{A}^{\mathrm{morph}}$, which encodes morphologically plausible ambiguity among the predefined defect classes (Section \ref{subsec:ambiguity_prior}). This matrix plays a central role in both training and inference. During training, $\mathbf{A}^{\mathrm{morph}}$ guides the model to allocate probability mass only to morphologically plausible alternatives instead of uniformly distributing it across all incorrect classes (Section \ref{subsec:training}). During inference, it determines whether the predicted ambiguity is sufficiently explained by known morphological relationships to justify assisted diagnosis; otherwise, the wafer is escalated for full review (Section \ref{subsec:inference}). We below describe each component in detail.

\subsubsection{Morphology Descriptors}
\label{subsec:morphology_descriptors}

\begin{figure}[t]
\centering
\includegraphics[width=1.0\linewidth]{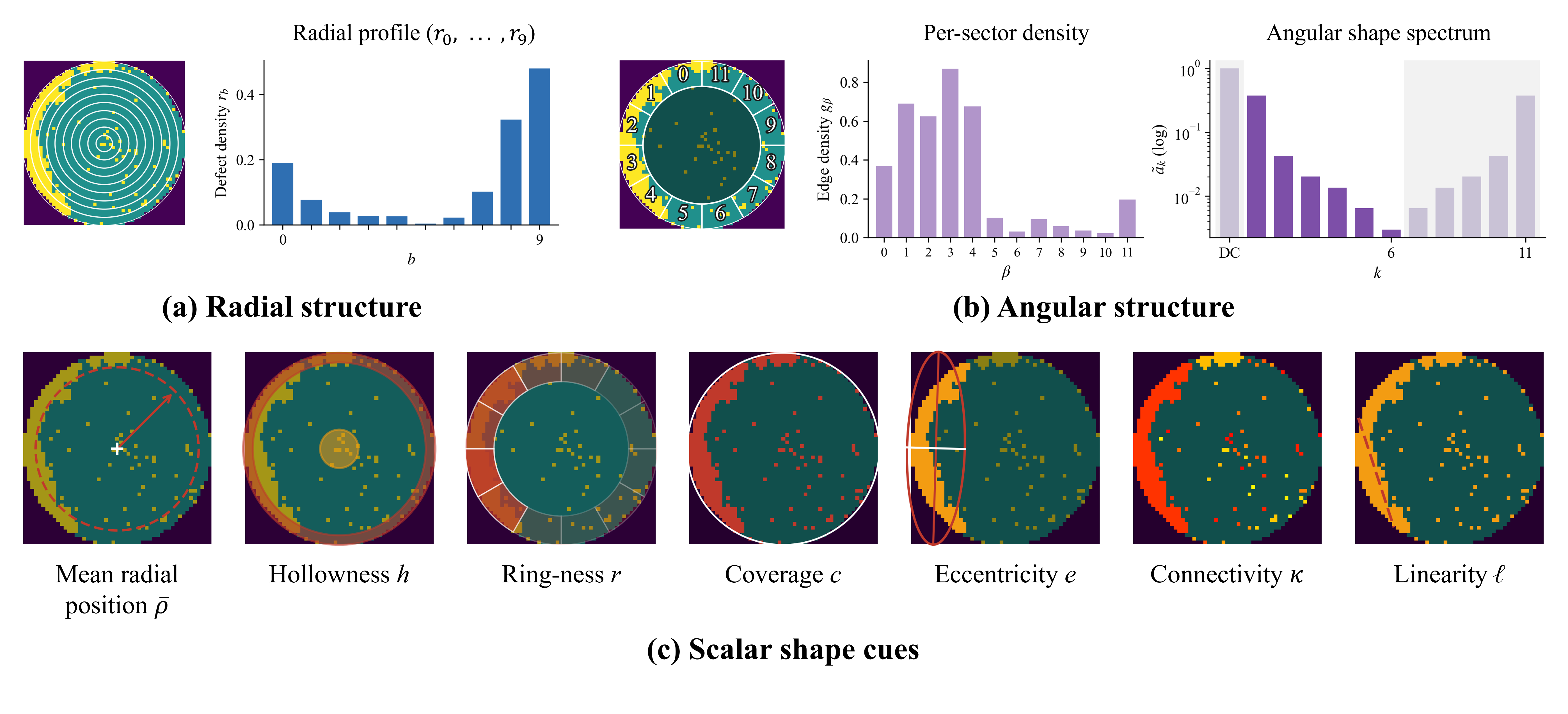}
\caption{The morphology descriptor $\boldsymbol{\phi}(x)$ on an Edge-Loc wafer. \textbf{(a)} Radial defect profile over ten concentric bins. \textbf{(b)} Defect density over twelve peripheral sectors and its rotation-invariant angular spectrum; shaded bars are not part of the descriptor. \textbf{(c)} The seven scalar shape cues, each with the geometry it measures overlaid.
}
\label{fig:morphology}
\end{figure}

For each wafer map $x$, we derive a deterministic 23-dimensional morphology descriptor $\boldsymbol{\phi}(x)$ from its wafer-support and defect masks. The descriptor consists of three complementary components: a 10-dimensional radial profile, a 6-dimensional rotation-invariant angular representation, and seven scalar shape features:
\begin{equation*}
\boldsymbol{\phi}(x)
=
\bigl[
\underbrace{\boldsymbol{\phi}_{\mathrm{rad}}(x)}_{10},
\underbrace{\boldsymbol{\phi}_{\mathrm{ang}}(x)}_{6},
\underbrace{\boldsymbol{\phi}_{\mathrm{shape}}(x)}_{7}
\bigr]^\top
\in \mathbb{R}^{23}.
\end{equation*}
As illustrated in Figure~\ref{fig:morphology}, these components describe where the defects occur radially, how they are distributed around the wafer, and what geometric form they exhibit. All features are computed directly from the wafer map without using a learned model. Below, we present the main ideas behind the descriptors, with mathematical and implementation details deferred to Supplementary Material~A.

\paragraph{Radial profile.}
We divide the wafer into ten concentric regions and record the defective-die density within each region. This profile captures whether the defects are concentrated near the wafer center, distributed around an intermediate annulus, or located close to the wafer boundary. It is therefore particularly useful for distinguishing patterns such as \texttt{Center}, \texttt{Donut}, and edge-related defects \cite{wu2015wafer}. We use ten radial regions to balance spatial resolution and profile stability. Given the typical wafer radius of 20--40 pixels, each region spans roughly 2--4 pixels, which is sufficient to distinguish major radial defect patterns without producing an overly fragmented profile.

\paragraph{Angular representation.}
To characterize how defects are distributed around the wafer, we divide the peripheral region into twelve sectors ($30^\circ$ each) and measure the defect density within each sector, following prior sector-based wafer-map analysis \cite{hu2026hybrid}. The resulting profile distinguishes defects distributed around the wafer circumference from those concentrated in specific angular regions. Because the absolute orientation of a defect pattern is not diagnostically meaningful on a circular wafer, we transform the sector profile using its Fourier spectrum to remove sensitivity to rotation. We further normalize the spectrum by its zero-frequency component, the direct-current (DC) term, which depends only on the mean sector density and thus removes the effect of overall defect density. The resulting descriptor thus captures the angular distribution of defects rather than their quantity or orientation.

\paragraph{Scalar shape features.}

In addition to the radial and angular representations, we use seven scalar features to characterize the location, extent, shape, and connectivity of defect patterns. These features are motivated by regional, geometric, clustering-based, and linear attributes used in prior wafer-map studies \cite{wu2015wafer,jin2019novel}.

\begin{itemize}

    \item  \textit{Mean radial position $\bar{\rho}$} measures the average normalized distance of defective dies from the wafer center. Small values indicate center-concentrated defects, whereas large values indicate defects located closer to the wafer boundary. This feature is analogous to the regional attributes of Wu \textit{et al.} \cite{wu2015wafer}, which measure the distance between the most salient defect region and the wafer center.

    \item  \textit{Hollowness $h$} measures how empty the wafer center is relative to the most densely occupied radial region. A high value indicates an annular pattern with a relatively empty center, helping distinguish  \texttt{Donut}-like patterns from filled \texttt{Center}-like patterns.

    \item  \textit{Ring-ness $r$} measures how uniformly defects are distributed around the wafer periphery. A high value indicates a uniformly occupied circumference, whereas a low value indicates that defects are concentrated within a limited angular region. It therefore helps distinguish \texttt{Edge-Ring} from \texttt{Edge-Loc}.

    \item \textit{Coverage ratio $c$} measures the proportion of wafer dies that are defective. It captures the overall extent of the defect pattern and is particularly informative for the \texttt{Near-full} class. Similar coverage-based descriptors have been used in prior wafer-map studies \cite{wu2015wafer,jin2019novel}.

    \item \textit{Eccentricity $e$} measures the elongation of the largest connected defect region. Values close to one indicate an elongated shape, whereas values close to zero indicate a more isotropic shape. It thus helps distinguish elongated \texttt{Scratch}-like patterns from more isotropic \texttt{Center}- and \texttt{Donut}-like patterns \cite{wu2015wafer}.

    \item \textit{Connectivity $\kappa$} measures how fragmented the defect pattern is. A high value indicates many small disconnected defect regions, as commonly observed in \texttt{Random}, whereas a low value indicates a more spatially coherent pattern, such as \texttt{Loc} or \texttt{Scratch}. This feature is motivated by clustering-based analyses of wafer defect patterns \cite{jin2019novel}.

    \item \textit{Linearity $\ell$} measures the extent to which defective dies form a dominant straight-line structure. A high value indicates a strongly linear pattern and is therefore characteristic of \texttt{Scratch}-like patterns. We derive this feature using a normalized Hough-transform response following the linear-attribute framework of \cite{wu2015wafer}.
\end{itemize}

To ensure that descriptor coordinates with different scales contribute comparably to the morphology-based characterization, we standardize each coordinate using statistics computed from the defect-class training samples, yielding $\widetilde{\boldsymbol{\phi}}(x)$. We exclude \texttt{Nonpattern} class from these statistics because its near-zero descriptors would dominate the scaling and suppress morphological variation among the defect classes. Coordinates with negligible variance are centered without scaling to avoid numerical instability. 

\subsubsection{Morphology-Derived Ambiguity Matrix}
\label{subsec:ambiguity_prior}

We construct a morphology-derived ambiguity matrix, $\mathbf{A}^{\mathrm{morph}}$, from the deterministic descriptor $\boldsymbol{\phi}(x)$ of each wafer map. The construction consists of three steps. First, we estimate a class-conditional distribution over the morphology descriptors of each defect class. Second, for each ordered pair of defect classes, we compute a directional morphology discrepancy that measures how well one class distribution explains samples from the other, and transform this discrepancy into a similarity score. Third, we row-normalize these similarities to obtain the ambiguity matrix. The entire procedure uses only training-set wafer maps and their labels. Accordingly, $\mathbf{A}^{\mathrm{morph}}$ is computed once before classifier training and is shared across all classifier backbones. It therefore represents morphology-derived class relations rather than a confusion matrix estimated from classifier errors.

\paragraph{Step 1: Estimating class-conditional morphology distributions.} 

Let $\widetilde{\boldsymbol{\phi}}(x) \in \mathbb{R}^{23}$ denote the standardized morphology descriptor of wafer $x$. For each defect class $j \in \mathcal{Y}_d$, we fit a diagonal-covariance Gaussian mixture model (GMM) to the descriptors of the training wafers assigned to class $j$:
\begin{equation*}
q_j(\boldsymbol{\varphi})
=
\sum_{m=1}^{M_j}
w_{jm}\,
\mathcal{N}\!\bigl(
\boldsymbol{\varphi};\,
\boldsymbol{\mu}_{jm},\,
\operatorname{diag}(\boldsymbol{\sigma}^2_{jm})
\bigr).
\end{equation*}
Here, $\boldsymbol{\varphi} \in \mathbb{R}^{23}$ denotes a generic standardized morphology descriptor, and $q_j(\boldsymbol{\varphi})$ is the estimated descriptor density for class $j$. Moreover, $M_j$ is the number of mixture components, $w_{jm}$ is the weight of component $m$, and $\boldsymbol{\mu}_{jm}$ and $\boldsymbol{\sigma}^2_{jm}$ denote its mean and diagonal variance, respectively. The fitting procedure is detailed in Supplementary Material~A.

\paragraph{Step 2: Calculating directional morphology similarity.}
For two defect classes $j,k \in \mathcal{Y}_d$, we quantify how well the morphology distribution of class $k$ explains wafers from class $j$ using the directional log-likelihood ratio
\begin{equation}
d(j \to k)
=
\mathbb{E}_{x:y=j}
\left[
\log q_j\!\bigl(\widetilde{\boldsymbol{\phi}}(x)\bigr)
-
\log q_k\!\bigl(\widetilde{\boldsymbol{\phi}}(x)\bigr)
\right].
\label{eq:morph_kl}
\end{equation}
Here, the expectation is taken over wafers assigned to class $j$.

Let $\mathcal{D}_{\mathrm{train}} =\{(x_n,y_n)\}_{n=1}^{N}$ denote the training set, and let $n_j = |\{n:y_n=j\}|$ be the number of training wafers in class $j$. We estimate the directional discrepancy directly from the class-$j$ training samples:
\begin{equation}
\widehat{d}(j \to k)
=
\frac{1}{n_j}
\sum_{n:y_n=j}
\log
\frac{
q_j\!\bigl(\widetilde{\boldsymbol{\phi}}(x_n)\bigr)
}{
q_k\!\bigl(\widetilde{\boldsymbol{\phi}}(x_n)\bigr)
}.
\label{eq:morph_discrepancy_est}
\end{equation}

A smaller value indicates that class $k$ explains the observed morphologies of class $j$ nearly as well as class $j$ itself. Because the expectation is taken over class-$j$ wafers, the resulting relation is directional: $d(j \to k)$ need not equal $d(k \to j)$. 

The directional discrepancy is converted into a similarity score as
\begin{equation*}
s(j \to k)
=
\exp\!\left(
-\frac{\widehat{d}_{\mathrm{shift}}(j \to k)}
{\tau_{\mathrm{sim}}}
\right),
\end{equation*}
where $\widehat{d}_{\mathrm{shift}}(j \to k)$ denotes the clipped and trimmed empirical discrepancy after shifting the minimum off-diagonal value to zero. We set $\tau_{\mathrm{sim}}$ to the median of the shifted off-diagonal discrepancies, so that the similarity scale is determined from the empirical discrepancy distribution. The exact procedure is provided in Supplementary Material~A.

\paragraph{Step 3: Constructing the ambiguity matrix.}
The ambiguity matrix $\mathbf{A}^{\mathrm{morph}} \in \mathbb{R}_{\geq 0}^{K \times K}$ separates the \texttt{Nonpattern} class from the eight named defect classes. Using the class indexing in Table~\ref{tab:class_index}, we define
\begin{equation}
\mathbf{A}^{\mathrm{morph}}
=
\begin{bmatrix}
1
&
\mathbf{0}^\top_{|\mathcal{Y}_d|}
\\
\mathbf{0}_{|\mathcal{Y}_d|}
&
\widetilde{\mathbf{A}}^{\mathrm{morph}}
\end{bmatrix},
\label{eq:amorph_block}
\end{equation}
where $\widetilde{\mathbf{A}}^{\mathrm{morph}} \in
\mathbb{R}_{\geq 0}^{|\mathcal{Y}_d| \times |\mathcal{Y}_d|}$ is a row-stochastic defect block. Note that the \texttt{Nonpattern} class is treated as unambiguous because it indicates the absence of an identifiable spatial defect pattern, rather than a specific defect morphology that could overlap with other defect classes.

For $j,k\in\mathcal{Y}_d$, the diagonal retains mass $\delta$ and the remaining mass $1-\delta$ is allocated across the alternative defect classes in proportion to the directional morphology similarities:
\begin{equation*}
\widetilde{A}^{\mathrm{morph}}_{jk}
=
\begin{cases}
\delta, & j=k, \\[6pt]
\displaystyle
(1-\delta)
\frac{s(j\to k)}
{\sum_{k'\neq j}s(j\to k')},
& j\neq k.
\end{cases}
\end{equation*}
where $\delta\in(0,1)$ controls the probability mass retained on the annotated class. The default value of $\delta$ is reported in Supplementary Material~A, and sensitivity to $\delta$ is examined in the ablation study in Section \ref{subsec:ablation}.

For later use, let us denote the $y-th$ row of $\mathbf{A}^{\mathrm{morph}}$ by
\begin{equation*}
\mathbf{a}_y
:=
\bigl[
A^{\mathrm{morph}}_{y,0},
A^{\mathrm{morph}}_{y,1},
\ldots,
A^{\mathrm{morph}}_{y,K-1}
\bigr]^\top
\in \Delta^{K-1},
\end{equation*}
where
$\Delta^{K-1}
=
\{\mathbf{v}\in\mathbb{R}_{\geq0}^{K}:
\sum_{i=0}^{K-1}v_i=1\}$
denotes the probability simplex. 
By construction, $\mathbf{a}_0$ is the one-hot vector for \texttt{Nonpattern}. For $y \in \mathcal{Y}_d$, its $y$-th entry is $\delta$, while the remaining weight (1-$\delta$) is distributed only among the other defect classes according to their morphological relationships. No weight is assigned to \texttt{Nonpattern}.

\subsubsection{Ambiguity-Aware Training}
\label{subsec:training}

We train the classifier by combining the standard hard-label loss with a soft-target loss whose targets are derived from the ambiguity matrix:
\begin{equation}
\mathcal{L}_{\mathrm{amb}}(\theta)
=
\mathbb{E}_{(x,y)}
\left[
(1-\lambda)\mathcal{L}_{\mathrm{hard}}(\theta;x,y)
+
\lambda\mathcal{L}_{\mathrm{soft}}(\theta;x,y)
\right],
\label{eq:amb_loss}
\end{equation}
where $\lambda \in [0,1]$ controls the contribution of ambiguity-aware supervision; setting $\lambda=0$ recovers standard cross-entropy training. We now define the two loss terms, $\mathcal{L}_{\mathrm{hard}}$ and $\mathcal{L}_{\mathrm{soft}}$.

Let $p_{\theta}(k\mid x)$ denote the classifier output for class $k$. For a training sample $(x,y)$, the hard-label loss is the usual cross-entropy,
\begin{equation*}
\mathcal{L}_{\mathrm{hard}}(\theta;x,y)
=
-\log p_{\theta}(y\mid x),
\end{equation*}
and the soft-target loss is the cross-entropy against a class-level target $\boldsymbol{\pi}_y$,
\begin{equation*}
\mathcal{L}_{\mathrm{soft}}(\theta;x,y)
=
-\sum_{k=0}^{K-1}
\pi_{y,k}\log p_{\theta}(k\mid x).
\end{equation*}

The target $\boldsymbol{\pi}_y$ is where the ambiguity matrix enters. For a training label $y \in \mathcal{Y}$,
\begin{equation}
\boldsymbol{\pi}_y
=
\begin{cases}
\mathbf{1}_y, & y=0, \\[4pt]
(1-\eta_y)\mathbf{1}_y+\eta_y \mathbf{a}_y,
& y \in \mathcal{Y}_d,
\end{cases}
\label{eq:soft_target}
\end{equation}
where $\mathbf{1}_y$ is the one-hot vector for class $y$, and $\eta_y = 1-\delta$ is the off-diagonal mass of $\mathbf{a}_y$, which controls the contribution of the morphology-derived matrix. Because the annotated class remains the dominant entry of $\mathbf{a}_y$, the target $\boldsymbol{\pi}_y$ preserves the original label while assigning a controlled amount of probability mass to morphologically plausible alternatives. The target is class-dependent rather than instance-dependent; all training wafers with label $y$ use the same $\boldsymbol{\pi}_y$.

\subsubsection{Selective Set-Valued Inference}
\label{subsec:inference}

At inference time, the ambiguity matrix determines whether a low-confidence prediction can be represented by a morphologically plausible two-class diagnostic set. Let $\widehat{y}_1$ and $\widehat{y}_2$ denote the top-1 and top-2 classes under $p_{\theta}(\cdot\mid x)$, and define the top-1 confidence as
\begin{equation*}
c_1
=
p_{\theta}(\widehat{y}_1\mid x).
\end{equation*}
The directional morphology plausibility of the top-2 pair is
\begin{equation*}
A_{\mathrm{pair}}(x)
=
A^{\mathrm{morph}}_{\widehat{y}_1,\widehat{y}_2}.
\end{equation*}
Given a confidence threshold $\tau_{\mathrm{conf}}$ and a morphology threshold $\tau_A$, the diagnostic rule is
\begin{equation}
\operatorname{decision}(x)
=
\begin{cases}
\{\widehat{y}_1\},
& c_1 \geq \tau_{\mathrm{conf}}, \\[4pt]
\{\widehat{y}_1,\widehat{y}_2\},
& c_1 < \tau_{\mathrm{conf}}
\ \text{and}\
A_{\mathrm{pair}}(x) \geq \tau_A, \\[4pt]
\textsc{full review},
& \text{otherwise}.
\end{cases}
\label{eq:decision_rule}
\end{equation}
The three outcomes correspond to automatic diagnosis, assisted diagnosis for morphologically interpretable boundary cases, and full review, respectively.
 Therefore, when the model has low confidence in its top prediction, it does not immediately escalate the wafer. Instead, it checks whether the two most probable classes form a morphologically plausible pair according to $\mathbf{A}^{\mathrm{morph}}$. If so, both classes are returned as a two-class diagnostic set; otherwise, the wafer is escalated for full review.

\subsection{Methodological Context and Design Choices}
\label{subsec:method_discussion}

Having presented the complete model, we now discuss the main design choices and clarify how the proposed framework differs from related alternatives.

\paragraph{Choice of the class-conditional density model.}
Density estimation in $\mathbb{R}^{23}$ requires a model that balances flexibility, statistical efficiency, and computational cost. Non-parametric kernel density estimators become unreliable in this dimensionality and incur query costs that grow with the number of training samples \cite{silverman1986density}. A single Gaussian per class cannot capture the within-class multimodality created when the same defect pattern appears at different wafer locations or orientations \cite{bishop2006prml}. More expressive parametric models, such as normalizing flows \cite{papamakarios2021normalizing}, require substantially more data and computation, which is inconsistent with the intended classifier-free, one-off construction of the matrix. A diagonal-covariance GMM provides a practical middle ground: it captures multimodality with a small number of components, supports efficient log-density evaluation in Eqs.~\eqref{eq:morph_kl}--\eqref{eq:morph_discrepancy_est}, and remains estimable for the smallest defect classes because its parameter count grows linearly with the descriptor dimension.

\paragraph{Relation to LS and label distribution learning.}
The proposed training objective is related to LS \cite{szegedy2016rethinking} and label distribution learning \cite{geng2016label,gao2017deep}, but differs in how the off-diagonal target mass is determined. Standard LS redistributes probability mass according to a fixed rule, typically uniformly across all classes. In contrast, Eq.~\eqref{eq:soft_target} allocates off-diagonal mass according to morphology-derived class relations. Label distribution learning generally assumes that a label distribution is observed or estimated for each instance, whereas the wafer-map dataset provides a single expert annotation for each wafer \cite{wu2015wafer}. The proposed target is therefore inferred from class-level morphology rather than from instance-level multi-label annotations.

In the experiments, the LS baseline uses standard cross-entropy with smoothing parameter $\varepsilon_{\mathrm{LS}}=0.1$, redistributing mass uniformly across all $K$ classes, including \texttt{Nonpattern}. This baseline evaluates whether generic confidence regularization is beneficial.

\paragraph{Relation to selective and set-valued prediction.}
The inference rule in Eq.~\eqref{eq:decision_rule} is related to reject-option and selective classification \cite{chow1970on,geifman2017selective,geifman2019SelectiveNet}, as well as set-valued and conformal prediction \cite{vovk2005Algorithmic,shafer2008atutorial}. Uncertainty-based selective and semi-automatic wafer-map classification has also been investigated \cite{alawieh2020wafer,yoon2022semi}. The proposed rule differs from confidence-only rejection because it does not send every low-confidence wafer directly to full review. Instead, it first determines whether the top-2 alternatives form a morphologically plausible pair.

The two-element set $\{\widehat{y}_1,\widehat{y}_2\}$ is itself a valid diagnostic output when the two candidates lead to the same downstream engineering action. When a single label is required, the set serves as a short list for focused human verification. The two-class branch therefore represents an operational diagnostic action rather than merely a statistical prediction set.

\paragraph{Structure-blind ambiguity control.}
To separate the benefit of morphology-specific structure from the effect of softening alone, we define a uniform ambiguity control $\mathbf{A}^{\mathrm{uniform}}$ with the same block structure as Eq.~\eqref{eq:amorph_block}. Its defect block is
\begin{equation*}
\widetilde{A}^{\mathrm{uniform}}_{jk}
=
\begin{cases}
\delta, & j=k, \\[4pt]
(1-\delta)/(|\mathcal{Y}_d|-1), & j \neq k.
\end{cases}
\end{equation*}
This control preserves the diagonal and total off-diagonal mass of $\mathbf{A}^{\mathrm{morph}}$ but removes all morphology-specific relations. Replacing $\mathbf{A}^{\mathrm{morph}}$ with $\mathbf{A}^{\mathrm{uniform}}$ in Eqs.~\eqref{eq:amb_loss}--\eqref{eq:soft_target} therefore yields a structure-blind ambiguity-aware training baseline. Comparing this baseline with the proposed method tests whether the learned morphology structure provides value beyond target softening itself.

\section{Experiments}
\label{sec:experiments}

In this section, we evaluate the proposed framework from the perspective of selective diagnosis rather than conventional top-1 classification. The experiments are organized around the following questions:

\begin{itemize}
    \item[\textbf{Q1.}]
    Does morphology-derived ambiguity structure maintain and improve classification performance and separate wafers into reliable automatic diagnoses, useful assisted diagnoses, and unresolved cases requiring full review?

    \item[\textbf{Q2.}]
    Does pair-specific morphology enable controllable and cost-efficient routing compared with structure-blind, confidence-based alternatives that share the same three-way action space?

    \item[\textbf{Q3.}]
    How sensitive is the proposed framework to the diagonal mass, and does its benefit transfer across classifier architectures?
\end{itemize}

\subsection{Common Experimental Protocol}
\label{subsec:exp_setup}

Experiments are conducted on the labeled portion of WM-811K \cite{wu2015wafer}, which contains one \texttt{Nonpattern} class and eight named defect-pattern classes. All main experiments use ResNet-34 \cite{he2016deep} as the reference backbone. For each random seed, the data are split into stratified training, validation, and test sets with a 70/10/20 ratio. The same split is shared by all methods within the same seed. Unless otherwise stated, results are reported as mean $\pm$ standard deviation over four different random seeds.

All wafer maps are resized to $224\times224$ using nearest-neighbor interpolation to preserve their discrete pixel semantics. Models use ImageNet-pretrained backbones adapted to single-channel inputs and are trained for 50 epochs with AdamW under the same optimization and augmentation protocol. Further implementation details are provided in Supplementary Material~B, and the exact definition of every reported metric in Supplementary Material~C.

Throughout the classification, routing, and sensitivity experiments, $\mathbf{A}^{\mathrm{uniform}}$ serves as a morphology-blind alternative for $\mathbf{A}^{\mathrm{morph}}$. It follows the same block structure as $\mathbf{A}^{\mathrm{morph}}$ in Eq.~\eqref{eq:amorph_block}, but distributes the off-diagonal mass uniformly across the defect classes:

\begin{equation*}
\widetilde{A}^{\mathrm{uniform}}_{jk}
=
\begin{cases}
\delta, & j=k, \\[4pt]
(1-\delta)/(|\mathcal{Y}_d|-1), & j\neq k.
\end{cases}
\end{equation*}

Thus, $\mathbf{A}^{\mathrm{uniform}}$ preserves the diagonal and total off-diagonal mass of $\mathbf{A}^{\mathrm{morph}}$ while removing morphological class relations.

Unless varied explicitly, ambiguity-aware variants use diagonal mass $\delta=0.8$ and soft-target mixing coefficient $\lambda=0.6$. The pair $(\delta,\lambda)$ was selected on the validation split over a grid of twelve combinations; the procedure and the full grid are reported in Supplementary Material~B. The selective inference thresholds are $\tau_{\mathrm{conf}}=0.95$ and $\tau_A=0.015$. The confidence threshold is deliberately strict, because an incorrect automatic diagnosis feeds the wrong failure mechanism into root-cause analysis and into the process adjustments that follow. Near-perfect accuracy, on the order of 99\% is generally expected before a wafer map classifier is deployed in production \cite{yoon2022semi}, and the same asymmetry motivates the reject option in selective classification \cite{chow1970on, alawieh2020wafer}.

\subsection{Main Results at the Default Operating Point}
\label{subsec:main}

This subsection answers \textbf{Q1} by examining classification behavior and three-way selective routing under the default configuration.

\subsubsection{Classification Performance and Structural Ablation}
\label{subsec:descriptor_ablation}

Before delving into the selective diagnostic behavior, we first examine whether ambiguity-aware training maintains strong conventional classification performance and whether the source of the ambiguity structure matters. Standard CE uses hard one-hot targets, whereas all-class LS \cite{szegedy2016rethinking} distributes a fixed amount of target mass uniformly across all nine classes.

In addition to $\mathbf{A}^{\mathrm{uniform}}$, we construct $\mathbf{A}^{\mathrm{pixel}}$ using the same pipeline as $\mathbf{A}^{\mathrm{morph}}$ but replacing the proposed descriptor with a 23-dimensional principal component analysis (PCA) projection of the raw $224\times224$ wafer-map pixels. PCA is fitted using defect-class training samples only. The resulting representation then undergoes the same downstream process as the morphology descriptor. This comparison allows us to examine whether the benefit of $\mathbf{A}^{\mathrm{morph}}$ stems specifically from the proposed morphology-aware representation $\boldsymbol{\phi}(\cdot)$, rather than simply from constructing a class-relation matrix extracted directly from raw pixel information.

Table \ref{tab:main_classification} reports macro-F1 over the full nine-class task. To focus on discrimination among the eight named defect patterns, we additionally report defect macro-F1 and defect balanced accuracy, both computed on the subset of test wafers whose annotation is a defect class. Restricting the evaluation to that subset isolates confusions \emph{among} defect morphologies from confusions between a defect and \texttt{Nonpattern}; consequently these values are not the column averages of the per-class scores reported later in Table~\ref{tab:per_class_f1}, which are computed on the full nine-class problem. Exact definitions are given in Supplementary Material~C.

\begin{table}[t]
\centering
\caption{Classification performance and structural ablation on WM-811K using ResNet-34.}
\label{tab:main_classification}
\label{tab:descriptor_source}
\small
\begin{tabular}{llccc}
\hline
Method & Target structure & Macro-F1 & Defect macro-F1 & Defect bal. acc. \\
\hline
CE & Hard one-hot & 0.9154 $\pm$ 0.0059 & 0.9176 $\pm$ 0.0049 & 0.8934 $\pm$ 0.0060 \\
LS & Uniform across all classes & 0.9109 $\pm$ 0.0063 & 0.9174 $\pm$ 0.0074 & 0.8914 $\pm$ 0.0142 \\
$\mathbf{A}^{\mathrm{uniform}}$ & Uniform across defect classes & 0.9093 $\pm$ 0.0090 & 0.9120 $\pm$ 0.0097 & 0.8863 $\pm$ 0.0112 \\
$\mathbf{A}^{\mathrm{pixel}}$ & Raw-pixel-derived & 0.9114 $\pm$ 0.0043 & 0.9138 $\pm$ 0.0038 & 0.8855 $\pm$ 0.0095 \\
\textbf{$\mathbf{A}^{\mathrm{morph}}$} & \textbf{Morphology-derived} & \textbf{0.9201 $\pm$ 0.0076} & \textbf{0.9245 $\pm$ 0.0082} & \textbf{0.9009 $\pm$ 0.0153} \\
\hline
\end{tabular}
\end{table}

From the table we see that our proposed approach $\mathbf{A}^{\mathrm{morph}}$ achieves the best performance across all three evaluation metrics. In particular, the improvements in the defect-focused metrics indicate that morphology-aware supervision enhances discrimination among the defect cases. These results support the effectiveness of the proposed morphology-derived structure not only for the selective diagnostic mechanism introduced below, but also for improving the generic classification capability of the backbone model.

Furthermore, the results demonstrate that the source of the class-relation structure is important. Although $\mathbf{A}^{\mathrm{pixel}}$ uses the same matrix-construction pipeline as $\mathbf{A}^{\mathrm{morph}}$, its performance remains close to that of the uniform control $\mathbf{A}^{\mathrm{uniform}}$ (0.9138 vs. 0.9120 in defect macro-F1). This suggests that simply introducing pair-specific relations from a generic data-derived representation is not sufficient. In contrast, the proposed morphology descriptor explicitly captures radial, angular, and geometric characteristics that are closely related to the definitions of the wafer defect classes, allowing the resulting matrix to encode more diagnostically meaningful inter-class relationships.

To further examine where the improvement of $\mathbf{A}^{\mathrm{morph}}$ arises, Table~\ref{tab:per_class_f1} reports the class-wise F1 scores. The proposed matrix improves performance for nearly all defect classes, with particularly pronounced gains for \texttt{Loc} and \texttt{Scratch}. \texttt{Loc} and \texttt{Scratch} exhibit distinctive morphological characteristics while also sharing partial structural similarity with neighboring defect categories. The result is consistent with the intended role of morphology-aware supervision; it provides the greatest benefit for classes whose decision boundaries can be better characterized through meaningful inter-class morphology relations.
 
\begin{table}[H]
\centering
\caption{Per-class F1 on ResNet-34. Samples are test-set counts, identical
across seeds because the split is stratified. $\Delta$F1 denotes the difference
between the two matrices.}
\label{tab:per_class_f1}
\small
\begin{tabular}{lrccc}
\hline
Class & Samples & $\mathbf{A}^{\mathrm{morph}}$ & $\mathbf{A}^{\mathrm{uniform}}$ & $\Delta$F1 \\
\hline
\texttt{Center} & 859 & 0.9512 $\pm$ 0.0015 & 0.9493 $\pm$ 0.0060 & +0.0020 \\
\texttt{Donut} & 111 & 0.8915 $\pm$ 0.0100 & 0.8781 $\pm$ 0.0307 & +0.0134 \\
\texttt{Edge-Loc} & 1,038 & 0.8777 $\pm$ 0.0154 & 0.8600 $\pm$ 0.0101 & +0.0177 \\
\texttt{Edge-Ring} & 1,936 & 0.9840 $\pm$ 0.0038 & 0.9801 $\pm$ 0.0049 & +0.0038 \\
\texttt{Loc} & 718 & \textbf{0.8431 $\pm$ 0.0085} & 0.8195 $\pm$ 0.0122 & \textbf{+0.0236} \\
\texttt{Near-full} & 30 & 0.9583 $\pm$ 0.0326 & 0.9546 $\pm$ 0.0348 & +0.0037 \\
\texttt{Random} & 173 & 0.9173 $\pm$ 0.0041 & 0.9101 $\pm$ 0.0120 & +0.0072 \\
\texttt{Scratch} & 239 & \textbf{0.8642 $\pm$ 0.0128} & 0.8395 $\pm$ 0.0183 & \textbf{+0.0247} \\
\texttt{Nonpattern} & 29,486 & 0.9932 $\pm$ 0.0005 & 0.9924 $\pm$ 0.0002 & +0.0008 \\
\hline
\end{tabular}
\end{table}

\subsubsection{Three-way Diagnostic Routing}


We now evaluate whether the model separates wafers according to the intended roles of the three diagnostic actions. Table~\ref{tab:bucket_metrics} reports the coverage and prediction performance within each routed group. Coverage denotes the proportion of wafers assigned to each action; top-1 accuracy and top-2 inclusion indicate whether the annotated class matches the leading prediction or appears among the two leading predictions, respectively. 

\begin{table}[H]
\centering
\caption{Three-way diagnostic routing under $\mathbf{A}^{\mathrm{morph}}$ at the default operating point $(\tau_{\mathrm{conf}},\tau_A)=(0.95,0.015)$ using ResNet-34.}
\label{tab:bucket_metrics}
\small
\begin{tabular}{lccc}
\hline
 & Automatic & Assisted & Full review \\
\hline
Coverage    & 94.2\%  & 1.9\%  & 3.8\%  \\
Top-1 acc.      & 99.75\% & 77.1\% & 74.1\% \\
Top-2 incl. & --      & 95.2\% & 96.1\% \\
\hline
\end{tabular}
\end{table}


The three groups support different forms of diagnostic output. Automatic diagnosis covers 94.2\% of the test wafers with 99.75\% top-1 accuracy, showing that a single label is reliable for the large majority of cases. In the assisted group, top-1 accuracy falls to 77.1\%, but top-2 inclusion rises to 95.2\%. The uncertainty in these cases is therefore largely concentrated between two alternatives rather than dispersed across the taxonomy.

Interestingly, the full-review group achieves slightly higher top-2 inclusion than the assisted group (96.1\% vs. 95.2\%). This raises an important question: if the annotated class is retained among the top two predictions so frequently, why are these wafers routed to full review rather than assisted diagnosis? Top-2 inclusion alone does not answer this question because it does not consider which classes form the returned pair. We therefore examine the class composition of the assisted and full-review groups in Table~\ref{tab:bucket_composition}.


\begin{table}[H]
\centering
\caption{Composition of the assisted and full-review groups by annotated class
($\mathbf{A}^{\mathrm{morph}}$, ResNet-34). Counts are seed
means rounded to the nearest integer; shares are the count divided by the group
total. Each entry is rounded independently, so a column need not sum exactly to
its total.}
\label{tab:bucket_composition}
\small
\setlength{\tabcolsep}{4pt}
\begin{tabular}{lcccccc}
\hline
& \multicolumn{3}{c}{Assisted set} & \multicolumn{3}{c}{Full review} \\
\cline{2-4}\cline{5-7}
Annotated class & $n$ & Share & Top-2 incl. & $n$ & Share & Top-2 incl. \\
\hline
\texttt{Nonpattern} & 21 & 3.2\% & 0.0\%$^{\dagger}$ & 782 & 58.7\% & 100.0\% \\
\texttt{Center}     & 66 & 9.9\% & 97.3\% & 78 & 5.8\% & 93.9\% \\
\texttt{Donut}      & 43 & 6.5\% & 96.1\% & 3 & 0.2\% & 47.5\%$^{\dagger}$ \\
\texttt{Edge-Loc}   & 229 & 34.2\% & 99.3\% & 222 & 16.7\% & 93.8\% \\
\texttt{Edge-Ring}  & 62 & 9.3\% & 99.6\% & 27 & 2.0\% & 88.2\%$^{\dagger}$ \\
\texttt{Loc}        & 177 & 26.5\% & 98.2\% & 156 & 11.7\% & 87.5\% \\
\texttt{Near-full}  & 6 & 1.0\% & 97.2\%$^{\dagger}$ & 0 & 0.0\% & -- \\
\texttt{Random}     & 27 & 4.1\% & 90.9\%$^{\dagger}$ & 12 & 0.9\% & 93.6\%$^{\dagger}$ \\
\texttt{Scratch}    & 36 & 5.3\% & 99.3\% & 52 & 3.9\% & 87.3\% \\
\hline
All & 669 & 100\% & 95.2\% & 1{,}332 & 100\% & 96.1\% \\
\textbf{Defect-annotated only} & \textbf{648} & \textbf{96.8\%} & \textbf{98.2\%} & \textbf{550} & \textbf{41.3\%} & \textbf{90.9\%} \\
\hline
\end{tabular}

\vspace{2pt}
{\footnotesize $^{\dagger}$ Fewer than thirty wafers per seed; reported for completeness and not statistically meaningful.}
\end{table}

The high top-2 inclusion of the full-review group is largely driven by \texttt{Nonpattern}, which accounts for 58.7\% of this group. For these wafers, \texttt{Nonpattern} is almost always among the two leading predictions, resulting in 100\% top-2 inclusion. Because these cases make up more than half of the full-review group, they substantially raise its overall top-2 inclusion to 96.1\%. 



However, this does not mean that these wafers are suitable for assisted two-class diagnosis. A pair such as {\texttt{Nonpattern}, \texttt{Edge-Loc}} does not provide two specific defect morphologies for focused verification. The proposed ambiguity matrix instead defines assisted diagnosis through morphological relationships among the eight named defect classes. Pairs containing \texttt{Nonpattern} therefore do not qualify for assisted diagnosis and are routed to full review.


This distinction becomes clearer when considering only wafers annotated with one of the eight named defect classes. Top-2 inclusion is then 98.2\% for the assisted group, compared with 90.9\% for the full-review group. Thus, for named defect patterns, the assisted branch more consistently retains the annotated class within its two-class diagnostic output.


\subsection{Operational Behavior and Alternative Selective Policies}
\label{subsec:opcost}

This subsection answers \textbf{Q2} by examining operating-point control, routing cost, and the comparison with confidence-based routing.

\subsubsection{Operating Point Control}

We examine whether the morphology-derived ambiguity matrix provides flexible control over the division between assisted diagnosis and full review. This is important in practice because the desired use of assisted diagnosis may depend on available review capacity or how selectively two-class outputs should be provided. Related production studies have similarly combined machine-learning-based quality monitoring with explicit inspection-capacity constraints \cite{shih2025sampling}. We therefore vary the morphology threshold $\tau_A$, which determines how strong the morphological relationship between the two leading predictions must be for a low-confidence wafer to receive assisted diagnosis, while fixing $\tau_{\mathrm{conf}}=0.95$.

\begin{figure}[H]
\centering
\includegraphics[width=0.6\linewidth]{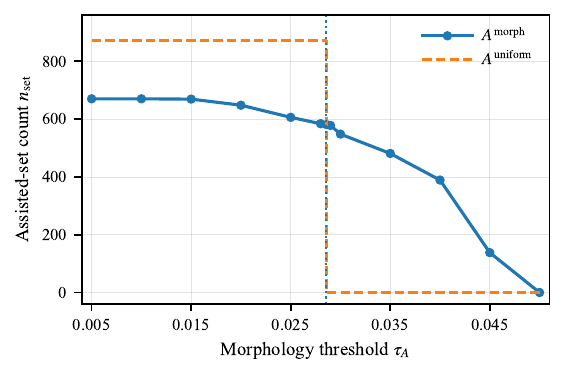}
\caption{Operating-point behavior as $\tau_A$ varies. The morphology-derived matrix produces a gradual change in assisted-set volume, whereas the uniform matrix exhibits a discrete cliff at its single off-diagonal value, $(1-\delta)/(|\mathcal{Y}_d|-1)=0.0286$.}
\label{fig:operating_curve}
\end{figure}

The morphology threshold applies only to low-confidence wafers whose two leading predictions are both among the eight named defect classes. Any pair containing \texttt{Nonpattern} has zero morphological plausibility and is therefore routed to full review for any positive $\tau_A$. Figure~\ref{fig:operating_curve} shows how the number of wafers assigned to assisted diagnosis changes with $\tau_A$, where $n_{\mathrm{set}}$ denotes the assisted-diagnosis group size.



Because $\mathbf{A}^{\mathrm{morph}}$ assigns different plausibility scores to different pairs of defect classes, increasing $\tau_A$ progressively moves pairs with weaker morphological support from assisted diagnosis to full review. The threshold therefore provides gradual control over the use of the intermediate diagnostic action. In contrast, all off-diagonal entries of $\mathbf{A}^{\mathrm{uniform}}$ are identical, so all defect-class pairs pass or fail the threshold together. As a result, its assisted group disappears at a single threshold rather than changing gradually. This comparison demonstrates a practical benefit of our method, enabling the assisted-diagnosis operating point to be adjusted according to different review requirements without abruptly eliminating the intermediate action.


\subsubsection{Routing Cost Analysis}
\label{subsubsec:costanalysis}

We next compare the routing policies using a stylized cost model. Related work has also examined similar inspection cost--yield trade-offs in semiconductor manufacturing
\cite{tirkel2016yield}. Here, full review has cost $1$, assisted verification has cost $\alpha$, and an incorrect automatic diagnosis has cost $10$. Although the coefficients do not estimate the absolute economic cost of a semiconductor fabrication workflow, the analysis provides a common set of relative penalties for examining how the routing policies allocate human effort and automatic risk. Table~\ref{tab:cost_analysis} reports the resulting expected routing costs under these relative penalties.
 
\begin{table}[H]
\centering
\caption{Expected per-sample routing cost on ResNet-34 under a common relative cost model. ``No assisted action'' removes the intermediate branch: a wafer that fails the confidence test is escalated rather than given a two-class set.}
\label{tab:cost_analysis}
\small
\begin{tabular}{llcccc}
\hline
Variant & $\tau_A$ & $\alpha=0.1$ & $\alpha=0.2$ & $\alpha=0.3$ & $\alpha=0.5$ \\
\hline
No assisted action & -- & 0.0810 & 0.0810 & 0.0810 & 0.0810 \\
\hline
$\mathbf{A}^{\mathrm{morph}}$ & 0.015 & \textbf{0.0636} & \textbf{0.0656} & \textbf{0.0675} & \textbf{0.0714} \\
$\mathbf{A}^{\mathrm{uniform}}$ & 0.015 & 0.0711 & 0.0737 & 0.0762 & 0.0812 \\
\hline
$\mathbf{A}^{\mathrm{morph}}$ & 0.040 & \textbf{0.0709} & \textbf{0.0720} & \textbf{0.0732} & \textbf{0.0754} \\
$\mathbf{A}^{\mathrm{uniform}}$ & 0.040 & 0.0938 & 0.0938 & 0.0938 & 0.0938 \\
\hline
\end{tabular}
\end{table}


Across all tested assisted-verification costs $\alpha$, morphology-aware routing achieves a lower cost than the two-action policy without assisted diagnosis. For example, at $\tau_A=0.015$ and $\alpha=0.2$, the cost decreases from 0.0810 to 0.0656, a 19\% reduction. Because both policies use the same confidence threshold for automatic diagnosis, this benefit comes from routing eligible low-confidence wafers to assisted diagnosis rather than full review.

The benefit remains at the stricter threshold $\tau_A=0.040$. At this threshold, the uniform ambiguity matrix produces no assisted diagnoses, whereas $\mathbf{A}^{\mathrm{morph}}$ retains strongly supported class pairs and maintains a lower cost than the two-action policy across all tested values of $\alpha$. This result highlights the benefit of pair-specific morphological relationships in selectively retaining the assisted action.


%

\subsubsection{Comparison with Simple Confidence-Based Routing}
\label{subsec:confidence_comparison}

We next compare our framework with a simple confidence-based routing. Our aim is to examine whether classifier confidence alone is sufficient to identify meaningful two-class diagnostic alternatives. For a direct comparison, both approaches use the same three diagnostic actions and the same confidence threshold, $\tau_{\mathrm{conf}}=0.95$. Let $p_1$ and $p_2$ denote the two highest predicted class probabilities of the CE-trained classifier. The confidence-based approach returns a single-class diagnosis when $p_1 \geq 0.95$, a two-class diagnostic set when $p_1<0.95$ but $p_1+p_2 \geq 0.95$, and otherwise sends the wafer to full review. Thus, unlike our framework, the baseline determines the two-class action solely from probability concentration, without considering the morphological relationship between the two classes. Table~\ref{tab:confidence_comparison} compares the resulting diagnostic actions.

\begin{table}[H]
\centering
\caption{Comparison of confidence-based routing and our framework. ``Named-patterns'' denotes the share of two-class outputs in which both classes correspond to named defect patterns, rather than one class being \texttt{Nonpattern}. Results are averaged over four seeds.}
\label{tab:confidence_comparison}
\small
\setlength{\tabcolsep}{4pt}
\begin{tabular}{lcccccc}
\hline
& \multicolumn{2}{c}{Automatic}
& \multicolumn{3}{c}{Assisted set}
& Full review \\
\cline{2-3}\cline{4-6}\cline{7-7}
Method & Coverage & Top-1 & Coverage & Named-patterns & Top-2 incl. & Coverage \\
\hline
CE $+$ confidence & 93.9\% & 99.76\% & 4.8\% & 24.8\% & 98.1\% & 1.2\% \\
Our framework     & 94.2\% & 99.75\% & 1.9\% & \textbf{100.0\%} & 95.2\% & 3.8\% \\
\hline
\end{tabular}
\end{table}

The main difference arises in assisted diagnosis. Confidence-based routing assigns 4.8\% of wafers to assisted diagnosis and only 1.2\% to full review, whereas our framework assigns 1.9\% and 3.8\%, respectively. However, the larger assisted coverage of confidence-based routing does not necessarily translate into more useful diagnostic support. Only 24.8\% of its two-class outputs contain two named defect patterns; most instead pair a defect class with \texttt{Nonpattern}, with common examples including \{\texttt{Nonpattern}, \texttt{Edge-Loc}\}, \{\texttt{Nonpattern}, \texttt{Loc}\}, and \{\texttt{Nonpattern}, \texttt{Center}\}. The frequent appearance of \texttt{Nonpattern} in assisted sets may result from its prevalence in the training data, which can make it more likely to receive high probability under standard CE training. Such outputs do not provide a choice between two specific defect morphologies and are thus less informative for assisted defect diagnosis. The higher top-2 inclusion of confidence-based routing (98.1\% vs. 95.2\%) should also be interpreted with caution, as pairs containing \texttt{Nonpattern} are counted as correct as long as they include the annotated class. In contrast, all two-class outputs from our framework consist of two named defect patterns. This demonstrates the advantage of incorporating morphological information: rather than expanding assisted coverage based solely on probability concentration, our framework selectively returns two-class diagnostic sets when the alternatives represent a meaningful morphological ambiguity.

Table~\ref{tab:assisted_examples} provides representative wafers that both approaches assign to assisted diagnosis. The examples illustrate the qualitative difference between confidence-based and morphology-aware routing. Although a defect pattern is visible in each wafer, confidence-based routing pairs the predicted defect with \texttt{Nonpattern}. Our framework instead returns a morphologically related defect alternative: \{\texttt{Scratch}, \texttt{Loc}\} for an elongated streak crossing the wafer, \{\texttt{Edge-Loc}, \texttt{Edge-Ring}\} for a partial arc along the wafer edge, and \{\texttt{Loc}, \texttt{Edge-Loc}\} for a local cluster extending toward the edge of the wafer. These three pairs are also the three most frequently confused defect pairs of the cross-entropy baseline, so the illustration reflects the dominant ambiguities of the task rather than isolated cases. These results show that probability concentration alone may be able to recognize uncertainty but does not necessarily make that uncertainty diagnostically interpretable. Incorporating morphology enables the framework to distinguish uncertain cases that admit a meaningful two-class diagnostic set from those that should instead be escalated for full review.

\begin{table}[t]
\centering
\caption{Representative wafers that both approaches route to assisted diagnosis.}
\label{tab:assisted_examples}
\small
\setlength{\tabcolsep}{5pt}
\renewcommand{\arraystretch}{1.05}
\begin{tabular}{@{}m{0.105\linewidth}m{0.105\linewidth}m{0.325\linewidth}m{0.325\linewidth}@{}}
\hline
Wafer map & True label & Confidence-based routing & Proposed approach \\
\hline
\includegraphics[width=\linewidth]{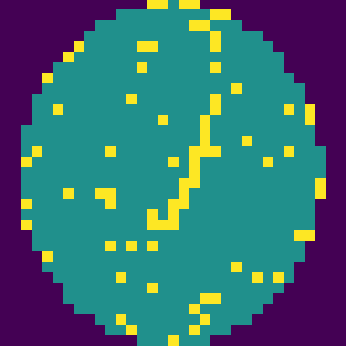} & \texttt{Scratch} & \{\texttt{Scratch}, \texttt{Nonpattern}\} & \{\texttt{Scratch}, \texttt{Loc}\} \\
\includegraphics[width=\linewidth]{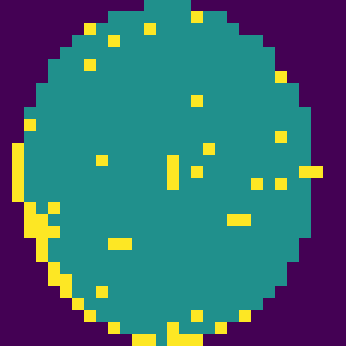} & \texttt{Edge-Loc} & \{\texttt{Edge-Loc}, \texttt{Nonpattern}\} & \{\texttt{Edge-Loc}, \texttt{Edge-Ring}\} \\
\includegraphics[width=\linewidth]{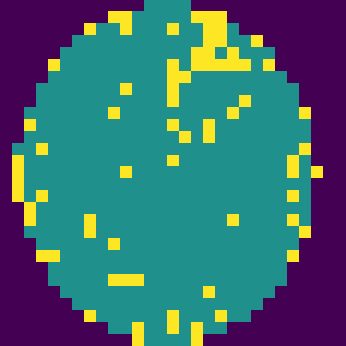} & \texttt{Loc} & \{\texttt{Loc}, \texttt{Nonpattern}\} & \{\texttt{Loc}, \texttt{Edge-Loc}\} \\
\hline
\end{tabular}
\end{table}

We also evaluated a variant of the CE-based approach in which any returned pair containing \texttt{Nonpattern} is discarded and the wafer escalated instead. Supplementary Material~D reports it. Under that variant the morphology gate admits 62\% more wafers to assisted diagnosis than the confidence gate.

\subsection{Sensitivity and Cross-backbone Robustness}
\label{subsec:ablation}

This subsection addresses \textbf{Q3} by examining the sensitivity of the proposed framework to the diagonal mass $\delta$ and its robustness to the choice of backbone architecture.

\subsubsection{Sensitivity to the Diagonal Mass}

The diagonal mass $\delta$ controls how strongly the model is encouraged to favor the annotated class relative to morphologically plausible alternatives during training. A small $\delta$ places greater emphasis on learning these alternatives. In contrast, a large $\delta$ makes it favor the annotated class, suppressing the influence of morphological relationships on learning. We selected $\delta$ on the validation split and report the corresponding test results in Table~\ref{tab:delta_ablation}, which also shows how sensitive the framework is to this trade-off.

\begin{table}[H]
\centering
\caption{Sensitivity to the diagonal mass $\delta$ using ResNet-34 at the selected mixing coefficient $\lambda=0.6$.}
\label{tab:delta_ablation}
\small
\begin{tabular}{clcc}
\hline
$\delta$
& Variant
& Macro-F1
& Defect macro-F1
\\
\hline
0.5 & $\mathbf{A}^{\mathrm{morph}}$ & 0.9161 $\pm$ 0.0096 & 0.9183 $\pm$ 0.0133 \\
0.5 & $\mathbf{A}^{\mathrm{uniform}}$ & 0.9189 $\pm$ 0.0043 & 0.9221 $\pm$ 0.0082 \\
0.8 & \textbf{$\mathbf{A}^{\mathrm{morph}}$} & \textbf{0.9201 $\pm$ 0.0076} & \textbf{0.9245 $\pm$ 0.0082} \\
0.8 & $\mathbf{A}^{\mathrm{uniform}}$ & 0.9093 $\pm$ 0.0090 & 0.9120 $\pm$ 0.0097 \\
0.9 & $\mathbf{A}^{\mathrm{morph}}$ & 0.9113 $\pm$ 0.0029 & 0.9148 $\pm$ 0.0051 \\
0.9 & $\mathbf{A}^{\mathrm{uniform}}$ & 0.9120 $\pm$ 0.0057 & 0.9150 $\pm$ 0.0042 \\
\hline
\end{tabular}
\end{table}

The results follow the expected trade-off. At $\delta=0.5$, the model places too much emphasis on alternative classes, weakening its ability to distinguish the annotated class from its alternatives; the morphology-derived matrix then falls slightly behind the uniform control on both metrics. At $\delta=0.9$, the model focuses more strongly on the annotated class, leaving less opportunity for morphological relationships to guide the learning of plausible alternatives, and the two matrices become indistinguishable. The intermediate value $\delta=0.8$ achieves the best balance between these two objectives, attaining the highest mean performance of any setting in the table on both metrics and the only clear margin over the uniform control.

\subsubsection{Cross-backbone Robustness}

We finally examine whether the proposed framework maintains consistent diagnostic behavior across different backbone architectures. We consider seven architectures: ResNet-18/34/50 \cite{he2016deep}, MobileNetV2 \cite{sandler2018mobilenetv2}, ShuffleNetV2 \cite{ma2018shufflenetv2}, EfficientNet-B0 \cite{tan2019efficientnet}, and Swin-Tiny \cite{liu2021swin}. The same $\mathbf{A}^{\mathrm{morph}}$ is applied without modification to every backbone. Table~\ref{tab:crossbackbone_routing} evaluates whether the three-way routing behavior remains consistent across backbone architectures.

\begin{table}[H]
\centering
\caption{Cross-backbone behavior of the selective diagnostic system under $\mathbf{A}^{\mathrm{morph}}$ at the default operating point $(\tau_{\mathrm{conf}},\tau_A)=(0.95,0.015)$. Each entry is a mean over four seeds; the last row is the standard deviation across the seven backbones.}
\label{tab:crossbackbone_routing}
\small
\setlength{\tabcolsep}{6pt}
\begin{tabular}{lcccc}
\hline
Backbone & Auto cov. & Auto top-1 & Asst. cov. & Asst. top-2 \\
\hline
ResNet-18 & 93.4\% & 99.80\% & 2.3\% & 95.7\% \\
ResNet-34 & 94.2\% & 99.75\% & 1.9\% & 95.2\% \\
ResNet-50 & 93.2\% & 99.77\% & 2.4\% & 95.9\% \\
MobileNetV2 & 93.1\% & 99.76\% & 2.3\% & 95.6\% \\
ShuffleNetV2 & 92.1\% & 99.79\% & 2.6\% & 95.8\% \\
EfficientNet-B0 & 93.2\% & 99.79\% & 2.1\% & 96.3\% \\
Swin-Tiny & 94.7\% & 99.72\% & 1.5\% & 94.6\% \\
\hline
Mean & 93.4\% & 99.77\% & 2.2\% & 95.6\% \\
SD & 0.8    & 0.03   & 0.4    & 0.6    \\
\hline
\end{tabular}
\end{table}

The framework shows consistent diagnostic behavior across all seven backbones. Automatic coverage ranges from 92.1\% to 94.7\%, while the accuracy of automatic diagnoses remains between 99.72\% and 99.80\%. Top-2 inclusion for assisted diagnosis is similarly stable, ranging from 94.6\% to 96.3\%. These results indicate that the division of wafers among automatic diagnosis, assisted diagnosis, and full review is largely insensitive to the choice of backbone. This versatility is practically valuable, as the framework can be paired with different backbone architectures according to deployment requirements, such as inference latency, memory constraints, or available computational resources. Notably, Swin-Tiny is not convolutional, so its consistent behavior shows that the framework does not depend on a convolutional inductive bias.

\section{Conclusion}
\label{sec:conclusion}

This paper reconsidered conventional single-label wafer map classification by recognizing that some wafers exhibit morphologies near the boundaries of known defect classes. To better support such cases, we formulated wafer defect diagnosis as a three-way decision: automatic single-class diagnosis, assisted diagnosis with two plausible defect classes, or full review. Central to the framework is a morphology-derived ambiguity matrix constructed from radial, angular, and geometric characteristics of training wafer maps. The matrix guides the model to learn plausible class alternatives during training and determines whether an uncertain prediction forms a meaningful two-class diagnostic set during inference.

Experiments on WM-811K demonstrate the benefits of incorporating morphological relationships into both learning and diagnostic routing. The proposed framework improves defect recognition over conventional and morphology-blind training approaches, provides more meaningful two-class alternatives than confidence-based routing, and reserves full review for cases with unresolved ambiguity. Scenario-based cost analyses further indicate the potential operational benefit of this three-way routing strategy. The diagnostic behavior of the framework remains consistent across various backbone architectures.

Several limitations also motivate future directions. First, the current ambiguity matrix represents class-level relationships and therefore cannot capture how ambiguity may vary across individual wafers within the same class pair. Future work could develop instance-adaptive ambiguity estimates to account for such variation. Second, WM-811K provides only a single annotation per wafer, preventing direct evaluation of whether engineers would regard both classes in a two-class diagnostic set as valid alternatives. Multi-annotator or engineer-validated datasets would enable stronger evaluation of this diagnostic interpretation. Third, our cost analysis relies on assumed relative costs for the different diagnostic actions rather than costs calibrated to an actual manufacturing workflow. Future studies could incorporate workflow-specific costs and evaluate the framework through engineer-in-the-loop deployment. Finally, the current framework focuses on boundary ambiguity within a fixed defect taxonomy. Extending it to mixed-type and previously unseen defect patterns remains an important direction for future work.

Despite these limitations, our findings demonstrate that wafer morphology can inform not only which defect class to predict, but also what level of diagnostic information should be provided. By distinguishing cases suitable for automatic diagnosis, assisted diagnosis, and full review, the proposed framework provides a natural basis for human-in-the-loop wafer classification. In such a system, clear cases can be handled automatically, interpretable ambiguity can be presented to engineers as focused diagnostic alternatives, and unresolved cases can be deferred for full review. Such adaptive allocation of diagnostic responsibility may enable more effective collaboration between automated classification systems and human expertise in semiconductor manufacturing.

\bibliographystyle{IEEEtran}
\bibliography{references}

@article{nakazawa2018wafer,
  title={Wafer Map Defect Pattern Classification and Image Retrieval Using Convolutional Neural Network},
  author={Nakazawa, Takeshi and Kulkarni, Deepak V.},
  journal={IEEE Transactions on Semiconductor Manufacturing},
  volume={31},
  number={2},
  pages={309--314},
  year={2018},
  month=may,
  doi={10.1109/TSM.2018.2795466},
  publisher={IEEE}
}

@article{wu2015wafer,
   author    = {Wu, Ming-Ju and Jang, Jyh-Shing R. and Chen, Jui-Long},
   title     = {Wafer Map Failure Pattern Recognition and Similarity Ranking for Large-Scale Data Sets},
   journal   = {IEEE Transactions on Semiconductor Manufacturing},
   volume    = {28},
   number    = {1},
   pages     = {1--12},
   year      = {2015},
   month     = feb,
   doi       = {10.1109/TSM.2014.2364237}
}

@inproceedings{he2016deep,
  author    = {He, Kaiming and Zhang, Xiangyu and Ren, Shaoqing and Sun, Jian},
  title     = {Deep Residual Learning for Image Recognition},
  booktitle = {Proceedings of the {IEEE} Conference on Computer Vision and Pattern Recognition ({CVPR})},
  pages     = {770--778},
  year      = {2016},
  doi       = {10.1109/CVPR.2016.90}
}

@inproceedings{sandler2018mobilenetv2,
  author    = {Sandler, Mark and Howard, Andrew and Zhu, Menglong and Zhmoginov, Andrey and Chen, Liang-Chieh},
  title     = {{MobileNetV2}: Inverted Residuals and Linear Bottlenecks},
  booktitle = {Proceedings of the {IEEE} Conference on Computer Vision and Pattern Recognition ({CVPR})},
  pages     = {4510--4520},
  year      = {2018},
  doi       = {10.1109/CVPR.2018.00474}
}

@inproceedings{tan2019efficientnet,
  author    = {Tan, Mingxing and Le, Quoc V.},
  title     = {{EfficientNet}: Rethinking Model Scaling for Convolutional Neural Networks},
  booktitle = {Proceedings of the 36th International Conference on Machine Learning ({ICML})},
  series    = {Proceedings of Machine Learning Research},
  volume    = {97},
  pages     = {6105--6114},
  year      = {2019},
  publisher = {PMLR}
}

@inproceedings{ma2018shufflenetv2,
  author    = {Ma, Ningning and Zhang, Xiangyu and Zheng, Hai-Tao and Sun, Jian},
  title     = {{ShuffleNet V2}: Practical Guidelines for Efficient {CNN} Architecture Design},
  booktitle = {Computer Vision -- {ECCV} 2018},
  editor    = {Ferrari, Vittorio and Hebert, Martial and Sminchisescu, Cristian and Weiss, Yair},
  series    = {Lecture Notes in Computer Science},
  volume    = {11218},
  pages     = {122--138},
  year      = {2018},
  publisher = {Springer},
  address   = {Cham},
  doi       = {10.1007/978-3-030-01264-9_8}
}

@inproceedings{liu2021swin,
  author    = {Liu, Ze and Lin, Yutong and Cao, Yue and Hu, Han and Wei, Yixuan and Zhang, Zheng and Lin, Stephen and Guo, Baining},
  title     = {{Swin Transformer}: Hierarchical Vision Transformer using Shifted Windows},
  booktitle = {Proceedings of the {IEEE/CVF} International Conference on Computer Vision ({ICCV})},
  pages     = {9992--10002},
  year      = {2021},
  doi       = {10.1109/ICCV48922.2021.00986}
}

@inproceedings{loshchilov2019decoupled,
  author    = {Loshchilov, Ilya and Hutter, Frank},
  title     = {Decoupled Weight Decay Regularization},
  booktitle = {International Conference on Learning Representations ({ICLR})},
  year      = {2019},
  eprint    = {1711.05101},
  archivePrefix = {arXiv}
}

@article{chen2000aneural,
  author  = {Chen, F. L. and Liu, S. F.},
  title   = {A Neural-Network Approach to Recognize Defect Spatial Pattern in Semiconductor Fabrication},
  journal = {IEEE Transactions on Semiconductor Manufacturing},
  year    = {2000},
  volume  = {13},
  number  = {3},
  pages   = {366--373},
  month   = aug,
  doi     = {10.1109/66.857947}
}

@article{yuan2011detection,
  author  = {Yuan, Tao and Kuo, Way and Bae, Suk Joo},
  title   = {Detection of Spatial Defect Patterns Generated in Semiconductor Fabrication Processes},
  journal = {IEEE Transactions on Semiconductor Manufacturing},
  year    = {2011},
  volume  = {24},
  number  = {3},
  pages   = {392--403},
  month   = aug,
  doi     = {10.1109/TSM.2011.2154870}
}

@article{yu2019wafer,
  author  = {Yu, Naigong and Xu, Qiao and Wang, Honglu},
  title   = {Wafer Defect Pattern Recognition and Analysis Based on Convolutional Neural Network},
  journal = {IEEE Transactions on Semiconductor Manufacturing},
  year    = {2019},
  volume  = {32},
  number  = {4},
  pages   = {566--573},
  month   = nov,
  doi     = {10.1109/TSM.2019.2937793}
}

@article{kang2020rotation,
  author  = {Kang, Seokho},
  title   = {Rotation-Invariant Wafer Map Pattern Classification With Convolutional Neural Networks},
  journal = {IEEE Access},
  year    = {2020},
  volume  = {8},
  pages   = {170650--170658},
  doi     = {10.1109/ACCESS.2020.3024603}
}

@article{kahng2021self,
  author  = {Kahng, Hyungu and Kim, Seoung Bum},
  title   = {Self-Supervised Representation Learning for Wafer Bin Map Defect Pattern Classification},
  journal = {IEEE Transactions on Semiconductor Manufacturing},
  year    = {2021},
  volume  = {34},
  number  = {1},
  pages   = {74--86},
  month   = feb,
  doi     = {10.1109/TSM.2020.3038165}
}

@inproceedings{alawieh2020wafer,
author={Alawieh, Mohamed Baker and Boning, Duane and Pan, David Z.},
  booktitle={2020 57th {ACM/IEEE} Design Automation Conference ({DAC})}, 
  title={Wafer Map Defect Patterns Classification using Deep Selective Learning}, 
  year={2020},
  volume={},
  number={},
  pages={1--6},
  doi={10.1109/DAC18072.2020.9218580}
}

@article{wang2019AdaBalGAN,
  author  = {Wang, Junliang and Yang, Zhengliang and Zhang, Jie and Zhang, Qihua and Chien, Wei-Ting Kary},
  title   = {{AdaBalGAN}: An Improved Generative Adversarial Network With Imbalanced Learning for Wafer Defective Pattern Recognition},
  journal = {IEEE Transactions on Semiconductor Manufacturing},
  year    = {2019},
  volume  = {32},
  number  = {3},
  pages   = {310--319},
  month   = aug,
  doi     = {10.1109/TSM.2019.2925361}
}

@inproceedings{szegedy2016rethinking,
  author    = {Szegedy, Christian and Vanhoucke, Vincent and Ioffe, Sergey and Shlens, Jonathon and Wojna, Zbigniew},
  title     = {Rethinking the Inception Architecture for Computer Vision},
  booktitle = {Proceedings of the {IEEE} Conference on Computer Vision and Pattern Recognition ({CVPR})},
  year      = {2016},
  pages     = {2818--2826},
  doi       = {10.1109/CVPR.2016.308}
}

@article{geng2016label,
  author  = {Geng, Xin},
  title   = {Label Distribution Learning},
  journal = {IEEE Transactions on Knowledge and Data Engineering},
  year    = {2016},
  volume  = {28},
  number  = {7},
  pages   = {1734--1748},
  month   = jul,
  doi     = {10.1109/TKDE.2016.2545658}
}

@article{gao2017deep,
  author  = {Gao, Bin-Bin and Xing, Chao and Xie, Chen-Wei and Wu, Jianxin and Geng, Xin},
  title   = {Deep Label Distribution Learning With Label Ambiguity},
  journal = {IEEE Transactions on Image Processing},
  year    = {2017},
  volume  = {26},
  number  = {6},
  pages   = {2825--2838},
  month   = jun,
  doi     = {10.1109/TIP.2017.2689998}
}

@article{chow1970on,
  author  = {Chow, C.},
  title   = {On Optimum Recognition Error and Reject Tradeoff},
  journal = {IEEE Transactions on Information Theory},
  year    = {1970},
  volume  = {16},
  number  = {1},
  pages   = {41--46},
  month   = jan,
  doi     = {10.1109/TIT.1970.1054406}
}

@inproceedings{geifman2017selective,
author = {Geifman, Yonatan and El-Yaniv, Ran},
title = {Selective classification for deep neural networks},
booktitle = {Advances in Neural Information Processing Systems 30},
pages = {4878--4887},
year = {2017},
isbn = {9781510860964},
publisher = {Curran Associates Inc.},
address = {Red Hook, NY, USA},
numpages = {10},
location = {Long Beach, California, USA},
series = {NIPS'17}
}

@inproceedings{geifman2019SelectiveNet,
  author    = {Geifman, Yonatan and El-Yaniv, Ran},
  title     = {{SelectiveNet}: A Deep Neural Network With an Integrated Reject Option},
  booktitle = {Proceedings of the 36th International Conference on Machine Learning},
  year      = {2019},
  volume    = {97},
  series    = {Proceedings of Machine Learning Research},
  pages     = {2151--2159},
  publisher = {PMLR}
}

@book{vovk2005Algorithmic,
  author    = {Vovk, Vladimir and Gammerman, Alexander and Shafer, Glenn},
  title     = {Algorithmic Learning in a Random World},
  publisher = {Springer},
  address   = {New York, NY, USA},
  year      = {2005},
  isbn      = {978-0-387-00152-4}
}

@article{shafer2008atutorial,
  author  = {Shafer, Glenn and Vovk, Vladimir},
  title   = {A Tutorial on Conformal Prediction},
  journal = {Journal of Machine Learning Research},
  year    = {2008},
  volume  = {9},
  pages   = {371--421}
}

@article{jin2019novel,
  author  = {Jin, Cheng-Hao and Na, Hyuk-Jun and Piao, Minghao and Pok, Gouchol and Ryu, Keun-Ho},
  title   = {A Novel {DBSCAN}-Based Defect Pattern Detection and Classification Framework for Wafer Bin Map},
  journal = {IEEE Transactions on Semiconductor Manufacturing},
  year    = {2019},
  volume  = {32},
  number  = {3},
  pages   = {286--292},
  month   = aug,
  doi     = {10.1109/TSM.2019.2916835},
  publisher = {IEEE}
}

@article{tsai2020light,
  author  = {Tsai, Tsung-Han and Lee, Yu-Chen},
  title   = {A Light-Weight Neural Network for Wafer Map Classification Based on Data Augmentation},
  journal = {IEEE Transactions on Semiconductor Manufacturing},
  year    = {2020},
  volume  = {33},
  number  = {4},
  pages   = {663--672},
  month   = nov,
  doi     = {10.1109/TSM.2020.3013004},
  publisher = {IEEE}
}

@article{jang2020support,
  author  = {Jang, Jaeyeon and Seo, Minkyung and Kim, Chang Ouk},
  title   = {Support Weighted Ensemble Model for Open Set Recognition of Wafer Map Defects},
  journal = {IEEE Transactions on Semiconductor Manufacturing},
  year    = {2020},
  volume  = {33},
  number  = {4},
  pages   = {635--643},
  month   = nov,
  doi     = {10.1109/TSM.2020.3012183},
  publisher = {IEEE}
}

@article{park2021discriminative,
  author  = {Park, Seyoung and Jang, Jaeyeon and Kim, Chang Ouk},
  title   = {Discriminative Feature Learning and Cluster-Based Defect Label Reconstruction for Reducing Uncertainty in Wafer Bin Map Labels},
  journal = {Journal of Intelligent Manufacturing},
  year    = {2021},
  volume  = {32},
  number  = {1},
  pages   = {251--263},
  doi     = {10.1007/s10845-020-01571-4},
  publisher = {Springer}
}

@article{lee2020semi,
  author  = {Lee, Hyuck and Kim, Heeyoung},
  title   = {Semi-Supervised Multi-Label Learning for Classification of Wafer Bin Maps With Mixed-Type Defect Patterns},
  journal = {IEEE Transactions on Semiconductor Manufacturing},
  year    = {2020},
  volume  = {33},
  number  = {4},
  pages   = {653--662},
  month   = nov,
  doi     = {10.1109/TSM.2020.3027431},
  publisher = {IEEE}
}

@article{kwak2023swaco,
  author  = {Kwak, Min Gu and Lee, Young Jae and Kim, Seoung Bum},
  title   = {{SWaCo}: Safe Wafer Bin Map Classification With Self-Supervised Contrastive Learning},
  journal = {IEEE Transactions on Semiconductor Manufacturing},
  year    = {2023},
  volume  = {36},
  number  = {3},
  pages   = {416--424},
  month   = aug,
  doi     = {10.1109/TSM.2023.3280891},
  publisher = {IEEE}
}

@article{yoon2022semi,
  author  = {Yoon, Suhee and Kang, Seokho},
  title   = {Semi-Automatic Wafer Map Pattern Classification With Convolutional Neural Networks},
  journal = {Computers \& Industrial Engineering},
  year    = {2022},
  volume  = {166},
  pages   = {107977},
  doi     = {10.1016/j.cie.2022.107977},
  publisher = {Elsevier}
}

@INPROCEEDINGS{deng2009imagenet,
  author={Deng, Jia and Dong, Wei and Socher, Richard and Li, Li-Jia and Kai Li and Li Fei-Fei},
  booktitle={2009 {IEEE} Conference on Computer Vision and Pattern Recognition}, 
  title={{ImageNet}: A large-scale hierarchical image database}, 
  year={2009},
  volume={},
  number={},
  pages={248--255},
  doi={10.1109/CVPR.2009.5206848}}

@book{silverman1986density,
  author    = {Silverman, B. W.},
  title     = {Density Estimation for Statistics and Data Analysis},
  publisher = {Chapman \& Hall},
  address   = {London},
  year      = {1986}
}

@book{bishop2006prml,
  author    = {Bishop, Christopher M.},
  title     = {Pattern Recognition and Machine Learning},
  publisher = {Springer},
  year      = {2006}
}

@article{papamakarios2021normalizing,
  author  = {Papamakarios, George and Nalisnick, Eric and Rezende, Danilo Jimenez and Mohamed, Shakir and Lakshminarayanan, Balaji},
  title   = {Normalizing Flows for Probabilistic Modeling and Inference},
  journal = {Journal of Machine Learning Research},
  volume  = {22},
  number  = {57},
  pages   = {1--64},
  year    = {2021}
}

@book{mclachlan2000finite,
  author    = {McLachlan, Geoffrey and Peel, David},
  title     = {Finite Mixture Models},
  publisher = {John Wiley \& Sons},
  address   = {New York},
  year      = {2000}
}

@article{chien2013online,
  author  = {Chien, Chen-Fu and Hsu, Shao-Chung and Chen, Ying-Jen},
  title   = {A system for online detection and classification of wafer bin map defect patterns for manufacturing intelligence},
  journal = {International Journal of Production Research},
  year    = {2013},
  volume  = {51},
  number  = {8},
  pages   = {2324--2338},
  doi     = {10.1080/00207543.2012.737943}
}

@article{kim2020uncertain,
  author  = {Kim, Byunghoon and Jeong, Young-Seon and Tong, Seung Hoon and Jeong, Myong K.},
  title   = {A generalised uncertain decision tree for defect classification of multiple wafer maps},
  journal = {International Journal of Production Research},
  year    = {2020},
  volume  = {58},
  number  = {9},
  pages   = {2805--2821},
  doi     = {10.1080/00207543.2019.1637035}
}

@article{tirkel2016yield,
  author  = {Tirkel, Israel and Rabinowitz, Gad and Price, David and Sutherland, Doug},
  title   = {Wafer fabrication yield learning and cost analysis based on in-line inspection},
  journal = {International Journal of Production Research},
  year    = {2016},
  volume  = {54},
  number  = {12},
  pages   = {3578--3590},
  doi     = {10.1080/00207543.2015.1106609}
}

@article{shih2025sampling,
  author  = {Shih, Ming-Sung and Chen, James C. and Chen, Tzu-Li and Chiang, Chih-Hsiung and Hsu, Ching-Lan},
  title   = {Machine-learning-based sampling inspection under {OQC} capacity for real-time quality monitoring in the {TFT-LCD} industry},
  journal = {International Journal of Production Research},
  year    = {2025},
  volume  = {63},
  number  = {6},
  pages   = {2090--2113},
  doi     = {10.1080/00207543.2024.2395389}
}

@article{liao2026evidential,
  author  = {Liao, Xinting and Zhang, Jie and Wang, Junliang and Lyu, Youlong and Zhong, Ray Y. and Zhang, Qihua},
  title   = {Evidential feature differentiated learning for trustworthy recognition of mixed-type defects in wafer maps},
  journal = {International Journal of Production Research},
  year    = {2026},
  volume  = {64},
  number  = {15},
  pages   = {6395--6415},
  doi     = {10.1080/00207543.2026.2623534}
}

@article{hu2026hybrid,
  author  = {Hu, Qiuhan},
  title   = {A hybrid deep and handcrafted feature learning approach for imbalanced wafer map defect classification},
  journal = {Scientific Reports},
  year    = {2026},
  note    = {Advance online publication},
  doi     = {10.1038/s41598-026-68884-x}
}

@article{saqlain2020deep,
  author  = {Saqlain, Muhammad and Abbas, Qasim and Lee, Jong Yun},
  title   = {A deep convolutional neural network for wafer defect identification on an imbalanced dataset in semiconductor manufacturing processes},
  journal = {IEEE Transactions on Semiconductor Manufacturing},
  year    = {2020},
  month   = aug,
  volume  = {33},
  number  = {3},
  pages   = {436--444},
  doi     = {10.1109/TSM.2020.2994357}
}

\newpage
\setcounter{table}{0}
\renewcommand{\thetable}{S\arabic{table}}

\setcounter{figure}{0}
\renewcommand{\thefigure}{S\arabic{figure}}

\setcounter{equation}{0}
\renewcommand{\theequation}{S\arabic{equation}}

\section*{Supplementary Material A: Morphology Descriptor and Ambiguity Matrix Construction Details}
\label{supp:ambiguity_details}

This section provides the exact mathematical definitions and implementation details for constructing the deterministic morphology descriptor $\phi(x)$ and the resulting morphology-derived ambiguity matrix $\mathbf{A}^{\mathrm{morph}}$.

\subsection*{A.1 Morphology Descriptor Construction}

For each wafer map $x\in\{0,1,2\}^{H\times W}$, we construct a
deterministic 23-dimensional morphology descriptor
\begin{equation*}
\phi(x)
=
\Bigl[
r_0,\ldots,r_9,\;
\tilde{a}_1,\ldots,\tilde{a}_6,\;
\bar{\rho},h,r,c,e,\kappa,\ell
\Bigr]
\in\mathbb{R}^{23}.
\end{equation*}
The descriptor contains a 10-dimensional radial profile, a
6-dimensional rotation-invariant angular representation, and seven scalar shape features. All components are computed directly from the wafer map and involve no learned model.

\paragraph{Wafer geometry and masks.}
We define the wafer-support and defect masks as
\begin{equation*}
M_w(x)=\mathbf{1}[x\neq0],
\qquad
M_d(x)=\mathbf{1}[x=2].
\end{equation*}
Their corresponding coordinate sets are
\begin{equation*}
\Omega_w
=
\{(u,v):M_w(u,v)=1\},
\qquad
\Omega_d
=
\{(u,v):M_d(u,v)=1\},
\end{equation*}
with cardinalities
\begin{equation*}
N_w=|\Omega_w|,
\qquad
N_d=|\Omega_d|.
\end{equation*}

The wafer centroid is
\begin{equation*}
(c_u,c_v)
=
\frac{1}{N_w}
\sum_{(u,v)\in\Omega_w}(u,v),
\end{equation*}
and the wafer radius is
\begin{equation*}
R
=
\max\left(
\max_{(u,v)\in\Omega_w}
\sqrt{(u-c_u)^2+(v-c_v)^2},\;
1
\right),
\end{equation*}
where the lower bound of one pixel guards against degenerate wafer supports. For each wafer-support location, we define its normalized radial position and polar angle as
\begin{equation*}
\rho(u,v)
=
\frac{\sqrt{(u-c_u)^2+(v-c_v)^2}}{R},
\qquad
\theta(u,v)
=
\operatorname{atan2}(v-c_v,u-c_u).
\end{equation*}
When a wafer contains fewer than three defective dies, we return the zero descriptor to ensure that all components are well-defined.

\paragraph{Radial profile.}
We partition the normalized radial interval $[0,1)$ into $B_\rho=10$ equal-width regions. For each radial bin
$b\in\{0,\ldots,B_\rho-1\}$, define
\begin{align*}
\mathcal{W}_b
&=
\left\{
(u,v)\in\Omega_w:
\left\lfloor B_\rho\rho(u,v)\right\rfloor=b
\right\},
\\
\mathcal{D}_b
&=
\left\{
(u,v)\in\Omega_d\cap\Omega_w:
\left\lfloor B_\rho\rho(u,v)\right\rfloor=b
\right\}.
\end{align*}
The radial defect density is
\begin{equation*}
r_b(x)
=
\frac{|\mathcal{D}_b|}
{\max(|\mathcal{W}_b|,1)},
\qquad
b=0,\ldots,9,
\end{equation*}
with $r_b(x)=0$ when $|\mathcal{W}_b|=0$. Because each bin is normalized by the number of wafer-support locations it contains, $r_b\in[0,1]$ is a density rather than a count, and is therefore insensitive to the differing wafer sizes present in WM-811K.

We use ten radial bins because typical WM-811K wafer radii are
approximately 20--40 pixels, corresponding to roughly 2--4 pixels per bin. This resolution retains the center, intermediate-annulus, and peripheral structures relevant to \texttt{Center}, \texttt{Donut}, and edge-related defect classes while avoiding excessively fragmented profiles for sparse defect patterns.

\paragraph{Angular representation.}
We define the peripheral annulus using the normalized radial threshold $\rho_{\mathrm{edge}}=0.7$:
\begin{equation*}
\mathcal{E}
=
\{(u,v)\in\Omega_w:\rho(u,v)\geq\rho_{\mathrm{edge}}\}.
\end{equation*}
The angular range $[-\pi,\pi)$ is divided into $B_\theta=12$ equal-width sectors. Let $\mathcal{E}_\beta$ denote the wafer-support locations in peripheral sector $\beta$. The defect density in that sector is
\begin{equation*}
g_\beta(x)
=
\frac{
\left|
\left\{
(u,v)\in\Omega_d\cap\mathcal{E}:
\theta(u,v)\in\text{$\beta$-th sector}
\right\}
\right|
}{
\max(|\mathcal{E}_\beta|,1)
},
\quad
\beta=0,\ldots,B_\theta-1.
\end{equation*}

We use twelve sectors, corresponding to a $30^\circ$ angular discretization, following prior hand-crafted wafer-map descriptors \cite{hu2026hybrid}. Let
\begin{equation}
a_k(x)
=
\left|
\sum_{\beta=0}^{B_\theta-1}
g_\beta(x)
\exp\left(
-\frac{2\pi i k\beta}{B_\theta}
\right)
\right|^2,
\qquad
k=0,\ldots,B_\theta-1,
\label{eq:supp_angular_spectrum}
\end{equation}
denote the power spectrum of the sector profile. Two properties of \eqref{eq:supp_angular_spectrum} determine which coefficients are retained.

First, the zero-frequency, or direct-current (DC), coefficient carries no angular information. Writing $\mu_g(x)=B_\theta^{-1}\sum_\beta g_\beta(x)$ for the mean sector density,
\begin{equation*}
a_0(x)
=
\Bigl(\sum_{\beta} g_\beta(x)\Bigr)^{2}
=
B_\theta^{2}\,\mu_g(x)^{2},
\end{equation*}
so $a_0$ is a deterministic function of how much of the periphery is defective, not of how that defect mass is arranged around the wafer. It is also strongly correlated with the outer radial bins $r_8,r_9$, which already encode peripheral defect density. We therefore divide the spectrum by $a_0$ and discard it, which additionally makes the representation invariant to a uniform rescaling of the sector profile.

Second, because $(g_0,\ldots,g_{B_\theta-1})$ is real-valued, its spectrum is Hermitian-symmetric,
\begin{equation*}
a_k(x)=a_{B_\theta-k}(x),
\qquad
k=1,\ldots,B_\theta-1,
\end{equation*}
so the coefficients above the Nyquist index duplicate those below it and carry no additional information.

Combining the two properties, the retained angular representation is
\begin{equation}
\tilde{a}_k(x)
=
\begin{cases}
0,
& a_0(x)<10^{-12},
\\[4pt]
\dfrac{a_k(x)}{a_0(x)},
& \text{otherwise},
\end{cases}
\qquad
k=1,\ldots,\frac{B_\theta}{2}=6,
\label{eq:supp_angular_normalized}
\end{equation}
giving six coefficients. Lower indices $k$ describe broad angular asymmetry and higher indices describe finer angular variation. A wafer rotation induces a cyclic shift of $(g_0,\ldots,g_{B_\theta-1})$, which leaves every $a_k$ and hence every $\tilde{a}_k$ unchanged, so the representation is rotation-invariant.

\paragraph{Scalar shape features.}
The remaining seven coordinates capture defect location, hollowness, peripheral uniformity, coverage, elongation, fragmentation, and linearity.

\textit{Mean radial position.}
The mean normalized radial position of defective dies is
\begin{equation*}
\bar{\rho}(x)
=
\frac{1}{N_d}
\sum_{(u,v)\in\Omega_d}
\rho(u,v).
\end{equation*}
Small values indicate center-concentrated defects, whereas large values indicate defects closer to the wafer boundary. This is a regional-feature analog of the distance-to-center attributes of \cite{wu2015wafer}.

\textit{Hollowness.}
Hollowness compares the density in the two innermost radial bins with the peak radial density:
\begin{equation*}
h(x)
=
\begin{cases}
0,
& \max_b r_b(x)<10^{-8},
\\[4pt]
\operatorname{clip}\left(
1-
\dfrac{[r_0(x)+r_1(x)]/2}
{\max_b r_b(x)},
0,1
\right),
& \text{otherwise}.
\end{cases}
\end{equation*}
Large values indicate an empty center relative to the most densely occupied annulus, as in \texttt{Donut}-like patterns.

\textit{Ring-ness.}
Let
\begin{equation*}
\mu_g(x)
=
\frac{1}{B_\theta}
\sum_{\beta=0}^{B_\theta-1}g_\beta(x),
\qquad
\sigma_g(x)
=
\left[
\frac{1}{B_\theta}
\sum_{\beta=0}^{B_\theta-1}
\bigl(g_\beta(x)-\mu_g(x)\bigr)^2
\right]^{1/2}.
\end{equation*}
Ring-ness is defined as
\begin{equation*}
r(x)
=
\begin{cases}
0,
& \mu_g(x)<10^{-8},
\\[4pt]
\operatorname{clip}\left(
1-\dfrac{\sigma_g(x)}{\mu_g(x)},
0,1
\right),
& \text{otherwise}.
\end{cases}
\end{equation*}
A uniformly occupied peripheral region yields high ring-ness, whereas angularly concentrated defects yield low ring-ness. Ring-ness and the angular coefficients $\tilde{a}_k$ are complementary: the former is a single second-moment summary of the sector profile, whereas the latter resolve the frequency at which the angular variation occurs.

\textit{Coverage ratio.}
The fraction of defective dies within the wafer support is
\begin{equation*}
c(x)
=
\frac{N_d}{N_w}.
\end{equation*}
Coverage-related quantities are commonly used as wafer-map descriptors \cite{wu2015wafer,jin2019novel} and are particularly informative for \texttt{Near-full}. Since the angular block is normalized by its DC term, $c$ is the coordinate through which overall defect extent enters the descriptor.

\textit{Eccentricity.}
Let $\mathcal{C}\subseteq\Omega_d$ be the largest 8-connected component of the defect mask. If $|\mathcal{C}|<2$, we set $e(x)=0$. Otherwise, let $(\mu_u,\mu_v)$ be the centroid of $\mathcal{C}$ and define its second-moment matrix as
\begin{equation*}
\boldsymbol{\Sigma}(x)
=
\frac{1}{|\mathcal{C}|-1}
\sum_{(u,v)\in\mathcal{C}}
\begin{pmatrix}
u-\mu_u\\
v-\mu_v
\end{pmatrix}
\begin{pmatrix}
u-\mu_u\\
v-\mu_v
\end{pmatrix}^{\!\top}.
\end{equation*}
Let $\lambda_1(x)\geq\lambda_2(x)\geq0$ denote its eigenvalues.
Eccentricity is
\begin{equation*}
e(x)
=
\begin{cases}
0,
& \lambda_1(x)<10^{-8},
\\[4pt]
\sqrt{1-\lambda_2(x)/\lambda_1(x)},
& \text{otherwise}.
\end{cases}
\end{equation*}
This shape-moment construction follows the geometry-attribute framework of \cite{wu2015wafer}. Values near one indicate elongated structures, whereas values near zero indicate more isotropic components.

\textit{Connectivity.}
Let $J(x)$ be the number of 8-connected components in the defect mask. Connectivity is defined as
\begin{equation*}
\kappa(x)
=
\frac{J(x)}{N_d}.
\end{equation*}
This normalization distinguishes fragmented patterns containing many small components from coherent patterns of comparable defect size and is motivated by clustering-based wafer-map analysis \cite{jin2019novel}.

\textit{Linearity.}
We apply a Hough line transform to $M_d(x)$ using 60 candidate
orientations uniformly distributed over $[-\pi/2,\pi/2)$, and retain peaks whose accumulator response exceeds the threshold $T_H=3$ under a minimum peak separation of one distance bin and one angle bin. Let $a_\star$ be the largest surviving accumulator response, which counts the defective dies lying on the strongest detected straight line. Linearity is
\begin{equation*}
\ell(x)
=
\begin{cases}
0,
& \text{if no accumulator response exceeds }T_H,
\\[4pt]
\operatorname{clip}\left(
\dfrac{a_\star}{\max(N_d,1)},
0,1
\right),
& \text{otherwise}.
\end{cases}
\end{equation*}
Normalizing by the number of defective dies makes $\ell$ the fraction of the defect set explained by a single straight line, so that $\ell\in[0,1]$ measures \emph{alignment} rather than absolute line length. This feature follows the linear-attribute framework of \cite{wu2015wafer}; it is large for \texttt{Scratch}-like patterns and small for non-linear patterns.

\paragraph{Standardization.}
Each descriptor coordinate is standardized using statistics computed only from defect-class training wafers. For coordinate
$d\in\{1,\ldots,23\}$, let $\mu_d$ and $\sigma_d$ be its mean and standard deviation over training samples satisfying $y_n\in\mathcal{Y}_d$. We define
\begin{equation}
\widetilde{\phi}_d(x)
=
\begin{cases}
\dfrac{\phi_d(x)-\mu_d}{\sigma_d},
& \sigma_d\geq10^{-6},
\\[8pt]
\phi_d(x)-\mu_d,
& \sigma_d<10^{-6}.
\end{cases}
\label{eq:supp_standardized_phi}
\end{equation}
The \texttt{Nonpattern} class is excluded when computing these statistics because it is the majority class and its morphology descriptors are typically close to zero. Including it would dominate the global feature scale and suppress variation among the defect classes used to construct the ambiguity matrix.

\subsection*{A.2 Class-Conditional Density Estimation}

For each defect class $j\in\mathcal{Y}_d$, we fit a diagonal-covariance Gaussian mixture model (GMM) \cite{mclachlan2000finite} to the standardized descriptors $\widetilde{\phi}(x_n)$ of the training wafers with
$y_n=j$:
\begin{equation*}
q_j(\boldsymbol{\varphi})
=
\sum_{m=1}^{M_j}
w_{jm}
\mathcal{N}
\left(
\boldsymbol{\varphi};
\boldsymbol{\mu}_{jm},
\operatorname{diag}(\boldsymbol{\sigma}^2_{jm})
\right).
\end{equation*}
Diagonal covariance matrices are used because the smallest defect classes contain too few samples to reliably estimate full $23\times 23$ covariance matrices. The number of mixture components is selected by a sample-size heuristic:
\begin{equation*}
M_j
=
\max\left(
1,
\min\left(M_{\max},\left\lfloor\frac{n_j}{s}\right\rfloor\right)
\right),
\end{equation*}
where $n_j$ is the number of training samples in class $j$, $M_{\max}=3$, and $s=100$ samples per component. The expectation-maximization (EM) algorithm is run for at most $200$ iterations per initialization, with three random initializations. The fitted model with the largest training log-likelihood is retained. A diagonal regularization term of $10^{-3}$ is added to each covariance estimate for numerical stability.

\subsection*{A.3 Directional Empirical Distance}

The directional distance from class $j$ to class $k$ is computed using the log-likelihood difference of class-$j$ samples under the two class-conditional densities:
\begin{equation*}
\Delta_{n}(j,k)
=
\log q_j(\widetilde{\phi}(x_n))
-
\log q_k(\widetilde{\phi}(x_n)),
\qquad
y_n=j.
\end{equation*}
To reduce the effect of extreme low-likelihood samples, each difference is clipped to a fixed interval:
\begin{equation}
\bar{\Delta}_{n}(j,k)
=
\left[
\Delta_n(j,k)
\right]_{-10}^{+50},
\label{eq:supp_clipped_delta}
\end{equation}
where
\begin{equation*}
[z]_{a}^{b}
=
\min(\max(z,a),b).
\end{equation*}
The directional empirical distance is then
\begin{equation}
\hat{d}(j\to k)
=
\operatorname{TrimMean}_{0.05}
\left(
\left\{
\bar{\Delta}_{n}(j,k):y_n=j
\right\}
\right),
\label{eq:supp_dhat}
\end{equation}
where $\operatorname{TrimMean}_{0.05}$ removes the smallest and largest 5\% of values before averaging.

Before clipping and trimming, this quantity coincides exactly with the difference of empirical Kullback--Leibler (KL) divergences,
\begin{equation*}
\mathbb{E}_{x_n:y_n=j}
\left[\Delta_n(j,k)\right]
=
\operatorname{KL}(\hat{P}_j\|q_k)
-
\operatorname{KL}(\hat{P}_j\|q_j),
\end{equation*}
where $\hat{P}_j$ is the empirical descriptor distribution of class $j$; the entropy of $\hat{P}_j$ cancels between the two KL terms. Trimming and clipping in (\ref{eq:supp_clipped_delta})--(\ref{eq:supp_dhat}) make $\hat{d}(j\to k)$ a robust variant of this identity. The distance is anchored so that $\hat{d}(j\to j)=0$. Directionality is retained because morphological inclusion may be asymmetric. For example, a local edge defect may be plausible under a broader edge-local distribution, but the reverse relation need not have the same strength.

\subsection*{A.4 Shifted Distance and Similarity Conversion}

Empirical KL-like estimates can occasionally be negative because of finite-sample effects and density-estimation error. We therefore shift the off-diagonal distance matrix so that the smallest valid off-diagonal entry becomes zero. Let
\begin{equation*}
\mathcal{V}
=
\{(j,k):j,k\in\mathcal{Y}_d,\ j\neq k,\ q_j \text{ and } q_k \text{ are fitted}\}.
\end{equation*}
Then
\begin{equation*}
d^\dagger(j\to k)
=
\hat{d}(j\to k)
-
\min_{(j',k')\in\mathcal{V}}
\hat{d}(j'\to k').
\end{equation*}
The similarity temperature is selected automatically from the empirical distance scale:
\begin{equation*}
\tau_{\mathrm{sim}}
=
\operatorname{median}
\left\{
d^\dagger(j\to k):(j,k)\in\mathcal{V}
\right\}.
\end{equation*}
The corresponding similarity is
\begin{equation*}
s(j\to k)
=
\exp\left(
-d^\dagger(j\to k)/\tau_{\mathrm{sim}}
\right).
\end{equation*}
This automatic choice avoids introducing a manually tuned similarity temperature.

\subsection*{A.5 Defect-Block Matrix Assembly}

The defect-block ambiguity matrix is assembled by allocating the off-diagonal mass of each row across the alternative defect classes in proportion to the morphological similarities.

\paragraph{Initial off-diagonal mass.}
For each defect class $j\in\mathcal{Y}_d$, we allocate the total off-diagonal mass $1-\delta$ across the other defect classes in proportion to the morphological similarity $s(j\to k)$:
\begin{equation}
u_{jk}
=
(1-\delta)
\dfrac{s(j\to k)}
{\sum_{k'\neq j,\,(j,k')\in\mathcal{V}}s(j\to k')},
\qquad
j\neq k,\ (j,k)\in\mathcal{V},
\label{eq:supp_ujk}
\end{equation}
so that $\sum_{k'\neq j} u_{jk'}=1-\delta$ by construction. The default diagonal mass is $\delta=0.8$, and sensitivity to $\delta$ is evaluated in the main experiments.

\paragraph{Defect block.}
Because $\sum_{k'\neq j}u_{jk'}=1-\delta$ holds by construction in \eqref{eq:supp_ujk}, the defect block follows directly, with the diagonal held at $\delta$:
\begin{equation*}
\widetilde{A}^{\mathrm{morph}}_{jk}
=
\begin{cases}
\delta, & j=k,\\[4pt]
u_{jk}, & j\neq k,
\end{cases}
\end{equation*}
so every row sums to one. With $\delta=0.8$, each realized off-diagonal entry of $\mathbf{A}^{\mathrm{morph}}$ lies in $[0.014,\,0.049]$, and the reported matrix is determined by the morphological similarities alone.

\subsection*{A.6 Embedding Into the Full Class Space}

The defect-block matrix is embedded into the full $K\times K$ ambiguity matrix as
\begin{equation}
A^{\mathrm{morph}}_{ij}
=
\begin{cases}
\widetilde{A}^{\mathrm{morph}}_{ij}, & i,j\in\mathcal{Y}_d,\\
1, & i=j=0,\\
0, & \text{otherwise}.
\end{cases}
\label{eq:supp_full_amorph}
\end{equation}
Thus, each row of $\mathbf{A}^{\mathrm{morph}}$ sums to one. No probability mass is assigned between \texttt{Nonpattern} and the defect block. This design reflects the operational role of \texttt{Nonpattern}: it denotes the absence of an identifiable failure pattern rather than a positive defect morphology with a canonical spatial template.

The embedding also determines the behavior of the selective rule on pairs involving \texttt{Nonpattern}. Since $A^{\mathrm{morph}}_{0k}=A^{\mathrm{morph}}_{j0}=0$ for all $j,k\in\mathcal{Y}_d$, any low-confidence wafer whose two leading predictions include \texttt{Nonpattern} receives pair plausibility $a(x)=0<\tau_A$ and is escalated for any positive threshold. The exclusion of detection-level uncertainty from assisted diagnosis is therefore a consequence of \eqref{eq:supp_full_amorph}, not a separate rule added at inference time.

\subsection*{A.7 Uniform Ambiguity Control}

To isolate the effect of morphology-specific off-diagonal structure, we define a uniform control matrix:
\begin{equation*}
A^{\mathrm{uniform}}_{ij}
=
\begin{cases}
\delta, & i=j,\ i\in\mathcal{Y}_d,\\
(1-\delta)/(|\mathcal{Y}_d|-1), & i\neq j,\ i,j\in\mathcal{Y}_d,\\
1, & i=j=0,\\
0, & \text{otherwise}.
\end{cases}
\end{equation*}
This matrix preserves the diagonal mass and total off-diagonal mass of $\mathbf{A}^{\mathrm{morph}}$, but removes morphology-specific class relations. Therefore, comparing $\mathbf{A}^{\mathrm{morph}}$ and $\mathbf{A}^{\mathrm{uniform}}$ tests whether the learned behavior is due to morphology-aware structure rather than softening alone. At $\delta=0.8$ its single off-diagonal value is $(1-\delta)/(|\mathcal{Y}_d|-1)=0.0286$, which is the location of the threshold cliff observed in the operating-point analysis of the main text.

\subsection*{A.8 Descriptor-Source Control}

The descriptor-source control $\mathbf{A}^{\mathrm{pixel}}$ replaces $\phi(\cdot)$ with a principal component analysis (PCA) projection of the raw wafer-map pixels while leaving every subsequent step unchanged. Each wafer map is rendered with the input pipeline of \S B.1, flattened to a $224\times224$ vector, and projected onto the leading $23$ principal components. The PCA basis is fitted on defect-class training wafers only, matching the standardization scope of \eqref{eq:supp_standardized_phi}. The projected features then undergo the same defect-only standardization, class-conditional GMM fitting (\S A.2), directional distance and similarity conversion (\S A.3--A.4), clipping, and diagonal allocation with $\delta=0.8$ (\S A.5--A.6).

The projection dimension is set to $23$ so that $\mathbf{A}^{\mathrm{pixel}}$ and $\mathbf{A}^{\mathrm{morph}}$ share the descriptor dimensionality, the density model, and the entire matrix-construction pipeline. The only difference between them is the representation from which the class-conditional distributions are estimated, which is what allows the comparison in the main text to attribute any performance difference to the representation rather than to dimensionality or pipeline choices.

\section*{Supplementary Material B: Implementation and Hyperparameter Details}
\label{supp:impl}

This section describes the implementation details used in the experiments. The proposed training objective and selective inference rule are architecture-agnostic; the following choices specify the experimental protocol used for evaluation on WM-811K.

\subsection*{B.1 Input Pipeline}

Each raw wafer map $x\in\{0,1,2\}^{H\times W}$ is converted into a single-channel grayscale image. Pixel values are multiplied by 127, producing intensities $\{0,127,254\}$ for outside-wafer, intact-chip, and defective-chip locations. The image is converted to grayscale mode and resized to $224\times 224$ using nearest-neighbor interpolation to preserve the discrete wafer-map semantics. The image is then divided by 255 so that the input tensor lies in $[0,1]$ and has shape $(1,224,224)$.

The morphology descriptor of \S A.1 is computed from the original wafer map, before resizing. Resizing is required only by the image backbones; the descriptor, the ambiguity matrix, and all quantities derived from them are therefore independent of the input resolution used for classification.

\subsection*{B.2 Backbone Architectures}

The reference model uses ResNet-34 \cite{he2016deep}. For cross-backbone evaluation, we additionally use ResNet-18, ResNet-50, MobileNetV2 \cite{sandler2018mobilenetv2}, EfficientNet-B0 \cite{tan2019efficientnet}, ShuffleNetV2 \cite{ma2018shufflenetv2}, and Swin-Tiny \cite{liu2021swin}. Each backbone is initialized with ImageNet-pretrained weights \cite{deng2009imagenet}.

Because wafer maps are single-channel inputs, the first convolutional layer of each image backbone is adapted from three input channels to one input channel. Let the original pretrained first-layer weights be
\begin{equation*}
W^{\mathrm{old}}\in\mathbb{R}^{C_{\mathrm{out}}\times 3\times k\times k}.
\end{equation*}
The new single-channel weights are initialized by averaging over the RGB channel dimension:
\begin{equation*}
W^{\mathrm{new}}_{c,1,p,q}
=
\frac{1}{3}
\sum_{i=1}^{3}
W^{\mathrm{old}}_{c,i,p,q}.
\end{equation*}
The kernel size, stride, padding, dilation, and bias configuration are otherwise preserved.

The original classification head is replaced with a task-specific head:
\begin{equation*}
\operatorname{Dropout}(0.1)
\rightarrow
\operatorname{Linear}(d_{\mathrm{feat}},K),
\end{equation*}
where $K=9$ and $d_{\mathrm{feat}}$ is the feature dimension of the backbone.

\subsection*{B.3 Data Splits and Random Seeds}

The labeled WM-811K samples are split into train, validation, and test sets using a stratified split by class. The split proportions are
\begin{equation*}
70\% \text{ train} \;/\; 10\% \text{ validation} \;/\; 20\% \text{ test}.
\end{equation*}
All main results are reported over four seeds:
\begin{equation*}
\{7,13,21,42\}.
\end{equation*}
The same seed is used for the data split, data-loader initialization, and model initialization. All methods compared in a given seed use the identical train, validation, and test split. Because the split is stratified, the per-class test counts are identical across seeds.

\subsection*{B.4 Training Protocol}

All models are trained with AdamW \cite{loshchilov2019decoupled}. Unless otherwise stated, the default training hyperparameters are summarized in Table~\ref{tab:supp_training_hparams}.

\begin{table}[H]
\centering
\caption{Default training hyperparameters.}
\label{tab:supp_training_hparams}
\begin{tabular}{lc}
\hline
Hyperparameter & Value \\
\hline
Optimizer & AdamW \\
Learning rate & $10^{-4}$ \\
Weight decay & $10^{-4}$ \\
Batch size & 64 \\
Maximum epochs & 50 \\
Learning-rate schedule & Cosine annealing \\
Minimum learning rate & $10^{-6}$ \\
Mixed precision & FP16 on CUDA \\
Gradient clipping & 1.0 \\
Early-stopping metric & Validation macro-F1 \\
Early-stopping patience & 8 epochs \\
\hline
\end{tabular}
\end{table}

The final model for each run is the checkpoint with the best validation macro-F1. Pixel-level or human-review information is not used in model selection; all selection is performed using the validation classification labels.

\subsection*{B.5 Data Augmentation}

Training-time data augmentation consists of the following transformations:
\begin{itemize}
    \item random horizontal flip with probability $0.5$,
    \item random vertical flip with probability $0.5$,
    \item random rotation sampled uniformly from $[0,360^\circ]$.
\end{itemize}
These transformations reflect the fact that many wafer map defect patterns are meaningful up to rotation or reflection, while preserving the discrete wafer-map structure through nearest-neighbor interpolation. Augmentation is applied to the classifier input only. It does not affect the morphology descriptor, which is extracted once from the unaugmented wafer map, consistent with the rotation invariance of the angular block established in \eqref{eq:supp_angular_normalized}.

\subsection*{B.6 Ambiguity-Aware Training Hyperparameters}

The default descriptor, ambiguity-matrix, and training hyperparameters are summarized in Table~\ref{tab:supp_ambiguity_hparams}.

\begin{table}[t]
\centering
\caption{Default descriptor and ambiguity-aware training hyperparameters.}
\label{tab:supp_ambiguity_hparams}
\begin{tabular}{lc}
\hline
Hyperparameter & Value \\
\hline
Radial bins $B_\rho$ & 10 \\
Angular sectors $B_\theta$ & 12 \\
Peripheral radius threshold $\rho_{\mathrm{edge}}$ & 0.7 \\
Hough accumulator threshold $T_H$ & 3 \\
Descriptor dimension & 23 \\
Diagonal mass $\delta$ & 0.8 \\
Loss mixing coefficient $\lambda$ & 0.6 \\
GMM maximum components $M_{\max}$ & 3 \\
Samples per component $s$ & 100 \\
GMM covariance regularization & $10^{-3}$ \\
GMM maximum EM iterations & 200 \\
GMM initializations & 3 \\
\hline
\end{tabular}
\end{table}

At the default $\delta=0.8$, the total off-diagonal mass in each defect-class row of $\mathbf{A}^{\mathrm{morph}}$ is $0.2$. The target mixing weight is the off-diagonal mass itself, $\eta_y=1-\delta$, so $\eta_y=0.2$ for every defect class and $\eta_y=0$ for \texttt{Nonpattern}.

\paragraph{Selection of $(\delta,\lambda)$.}
The diagonal mass and the loss mixing coefficient were selected together on the validation split over the grid $\delta\in\{0.5,0.65,0.8,0.9\}\times\lambda\in\{0.3,0.6,1.0\}$, holding every other setting at the values above. The selection metric is validation defect macro-F1, computed over the eight defect classes on defect-annotated wafers only, which is the same definition used for the defect-focused metrics of the main text. Table~\ref{tab:supp_delta_lambda_grid} reports the full grid. The selected setting $(\delta,\lambda)=(0.8,0.6)$ attains the highest validation score of the twelve combinations. Test data were not used at any point in this selection.

\begin{table}[H]
\centering
\caption{Validation defect macro-F1 over the $(\delta,\lambda)$ grid, ResNet-34 with $\mathbf{A}^{\mathrm{morph}}$. Each entry is a mean over four seeds; the selected setting is shown in bold.}
\label{tab:supp_delta_lambda_grid}
\small
\begin{tabular}{cccc}
\hline
$\delta$ & $\lambda=0.3$ & $\lambda=0.6$ & $\lambda=1.0$ \\
\hline
$0.5$  & 0.9227 & 0.9226 & 0.9153 \\
$0.65$ & 0.9229 & 0.9252 & 0.9229 \\
$0.8$  & 0.9239 & \textbf{0.9314} & 0.9210 \\
$0.9$  & 0.9266 & 0.9257 & 0.9234 \\
\hline
\end{tabular}
\end{table}

The grid also shows why the two parameters have to be chosen jointly rather than one at a time. At $\lambda=0.3$ the best diagonal mass is $\delta=0.9$, whereas at the selected $\lambda=0.6$ it is $\delta=0.8$; sweeping $\delta$ at a fixed $\lambda=0.3$ would therefore have selected a different value.

\subsection*{B.7 Selective Inference Hyperparameters}

The default selective inference thresholds are
\begin{equation*}
(\tau_{\mathrm{conf}},\tau_A)=(0.95,0.015).
\end{equation*}
The confidence threshold $\tau_{\mathrm{conf}}$ controls entry into automatic diagnosis. The morphology threshold $\tau_A$ controls whether a low-confidence top-2 pair is accepted as an two-class diagnostic set. Both thresholds are held fixed for every backbone, seed, and ambiguity matrix. The value of $\tau_{\mathrm{conf}}$ reflects the near-perfect accuracy expected of a wafer map classifier before deployment, as discussed in the main text, and sensitivity to $\tau_A$ is analyzed in the operating-point sweep.

Two properties of the default $\tau_A$ are worth stating explicitly. First, it lies just above the smallest realized off-diagonal entries of $\mathbf{A}^{\mathrm{morph}}$ (\S A.5): two to three of the $56$ off-diagonal entries fall below it, depending on the seed, so at the default operating point the morphology gate rejects only a small fraction of the low-confidence defect--defect pairs; the sweep in the main text traverses the range over which the gate becomes progressively more selective. Second, because $\tau_A>0$, the embedding \eqref{eq:supp_full_amorph} sends every pair containing \texttt{Nonpattern} to full review irrespective of the threshold value.

\subsection*{B.8 Baselines and Controls}

The main baselines are standard cross-entropy (CE), all-class label smoothing, the uniform ambiguity matrix $\mathbf{A}^{\mathrm{uniform}}$, the descriptor-source control $\mathbf{A}^{\mathrm{pixel}}$, confidence-only rejection, and CE training combined with confidence-based three-way routing. The all-class label-smoothing baseline uses a smoothing parameter
\begin{equation*}
\varepsilon_{\mathrm{LS}}=0.1,
\end{equation*}
redistributing $\varepsilon_{\mathrm{LS}}$ of the target mass uniformly over all $K=9$ classes including \texttt{Nonpattern}. This is structurally distinct from $\mathbf{A}^{\mathrm{uniform}}$, which redistributes mass uniformly over the eight defect classes only and keeps the \texttt{Nonpattern} row hard. The two baselines therefore answer different questions: label smoothing asks whether generic confidence regularization helps, whereas $\mathbf{A}^{\mathrm{uniform}}$ asks whether structured ambiguity training restricted to the same scope as $\mathbf{A}^{\mathrm{morph}}$ helps without the morphological signal.

Confidence-only rejection uses the same confidence threshold $\tau_{\mathrm{conf}}$ but sends all non-automatic wafers directly to full review. It therefore has only two actions, automatic diagnosis and escalation, and serves as the reference point for the routing-cost analysis. It is evaluated on the same trained models as the proposed rule, so that comparison isolates the routing policy.

CE training with confidence-based three-way routing is the baseline that shares the proposed action space exactly. The classifier is trained with cross-entropy, and the routing rule reads only its output probabilities. Writing $p_1$ and $p_2$ for the two largest class probabilities, a wafer is diagnosed automatically when $p_1 \geq \tau_{\mathrm{conf}}$, receives the two-class set $\{\widehat{y}_1,\widehat{y}_2\}$ when $p_1 < \tau_{\mathrm{conf}}$ and $p_1+p_2 \geq \tau_{\mathrm{conf}}$, and is escalated otherwise. The rule introduces no new hyperparameter: it reuses $\tau_{\mathrm{conf}}=0.95$, and asks whether the classifier's belief is concentrated on the two leading classes. Because this baseline produces the same three actions as the proposed framework, the comparison isolates the criterion that decides which low-confidence wafers deserve an assisted diagnosis. Supplementary Material~D reports a further variant of it in which any pair containing \texttt{Nonpattern} is discarded.

\subsection*{B.9 Computational Cost of Ambiguity-Matrix Construction}

The morphology descriptor extraction, GMM fitting, and ambiguity-matrix assembly are performed once before classifier training. The resulting $K\times K$ matrix is reused for all downstream models, random seeds, and baselines that require the ambiguity matrix. Using CPU-only execution on an AMD Ryzen 9 9950X processor, the full preprocessing pipeline takes under five minutes for the WM-811K training set. This cost is negligible compared with repeated neural-network training.

At inference time the matrix contributes a single table lookup $A^{\mathrm{morph}}_{\widehat{y}_1,\widehat{y}_2}$ per wafer, performed only for wafers that fail the confidence test. The selective rule therefore adds no measurable overhead to classifier throughput.

\section*{Supplementary Material C: Evaluation Metrics and Protocols}
\label{supp:metrics}

This section states the exact definition of every reported quantity. Let $\mathcal{T}=\{(x_n,y_n)\}_{n=1}^{N_{\mathrm{test}}}$ denote the test split of one seed, and let
\begin{equation*}
\mathcal{T}_d=\{(x,y)\in\mathcal{T}:y\in\mathcal{Y}_d\}
\end{equation*}
denote the subset of test wafers whose annotation is one of the eight named defect classes.

\subsection*{C.1 Classification Metrics}

For a labeled collection $\mathcal{S}$ and a class $j$, let $\mathrm{F1}_j(\mathcal{S})$ denote the $F_1$ score of class $j$ computed from the precision and recall of the predictions on $\mathcal{S}$, with the convention $\mathrm{F1}_j=0$ when class $j$ has neither predicted nor annotated instances in $\mathcal{S}$.

\paragraph{Macro-F1.}
The overall macro-F1 is the unweighted mean over all $K=9$ classes, evaluated on the full test split:
\begin{equation*}
\mathrm{F1}^{\mathrm{macro}}
=
\frac{1}{K}\sum_{j=0}^{K-1}\mathrm{F1}_j(\mathcal{T}).
\end{equation*}

\paragraph{Defect macro-F1.}
The defect macro-F1 is the unweighted mean over the eight defect classes, evaluated on the defect-annotated subset $\mathcal{T}_d$:
\begin{equation}
\mathrm{F1}^{\mathrm{def}}
=
\frac{1}{|\mathcal{Y}_d|}\sum_{j\in\mathcal{Y}_d}\mathrm{F1}_j(\mathcal{T}_d).
\label{eq:supp_defect_f1}
\end{equation}
Two aspects of \eqref{eq:supp_defect_f1} require emphasis. First, the average runs over the eight defect classes by construction, not over the set of labels that happen to occur in $\mathcal{T}_d$ and its predictions. A defect wafer predicted as \texttt{Nonpattern} still counts as a false negative for its annotated class, but \texttt{Nonpattern} itself does not enter the average as a ninth term. Second, because $\mathcal{T}_d$ excludes \texttt{Nonpattern}-annotated wafers, a \texttt{Nonpattern} wafer predicted as a defect does not contribute a false positive to that defect class.

Consequently $\mathrm{F1}^{\mathrm{def}}$ is not equal to the average of the eight defect entries of the per-class table in the main text, which are computed on the full nine-class problem $\mathcal{T}$. The former isolates discrimination \emph{among} defect patterns; the latter additionally charges each defect class for its confusions with \texttt{Nonpattern}. Both quantities are reported, and neither is derivable from the other.

\paragraph{Defect balanced accuracy.}
The defect balanced accuracy is the unweighted mean recall over the eight defect classes on the same subset:
\begin{equation*}
\mathrm{BA}^{\mathrm{def}}
=
\frac{1}{|\mathcal{Y}_d|}
\sum_{j\in\mathcal{Y}_d}
\frac{\bigl|\{(x,y)\in\mathcal{T}_d:y=j,\ \widehat{y}_1(x)=j\}\bigr|}
{\bigl|\{(x,y)\in\mathcal{T}_d:y=j\}\bigr|}.
\end{equation*}

\subsection*{C.2 Routing Metrics}

The decision rule of the main text partitions $\mathcal{T}$ into three disjoint groups: the automatic group $\mathcal{A}$, the assisted group $\mathcal{Q}$, and the full-review group $\mathcal{F}$. For a group $\mathcal{G}\in\{\mathcal{A},\mathcal{Q},\mathcal{F}\}$ we report
\begin{align}
\text{coverage}(\mathcal{G})
&=
|\mathcal{G}|/N_{\mathrm{test}},
\nonumber
\\
\text{top-1}(\mathcal{G})
&=
\frac{1}{|\mathcal{G}|}
\sum_{(x,y)\in\mathcal{G}}
\mathbf{1}\bigl[\widehat{y}_1(x)=y\bigr],
\nonumber
\\
\text{top-2 incl.}(\mathcal{G})
&=
\frac{1}{|\mathcal{G}|}
\sum_{(x,y)\in\mathcal{G}}
\mathbf{1}\bigl[y\in\{\widehat{y}_1(x),\widehat{y}_2(x)\}\bigr].
\label{eq:supp_bucket_top2}
\end{align}
Top-2 inclusion is aligned with the deployed action only for the assisted group, where a two-element set is actually returned. For the automatic group the deployed output is $\{\widehat{y}_1\}$, and for the full-review group no automated label is issued at all; the top-1 and top-2 statistics of those groups are reported solely to characterize their composition.

Because top-2 inclusion is evaluated against the annotated class, it is degenerate for any wafer annotated \texttt{Nonpattern} that was escalated because \texttt{Nonpattern} appeared in its leading pair: such a wafer satisfies $y\in\{\widehat{y}_1,\widehat{y}_2\}$ by construction. Group-level top-2 inclusion for the full-review group is therefore reported alongside its decomposition by annotated class in the main text.

\subsection*{C.3 Routing Cost Model}

The stylized routing cost assigns $c_{\mathrm{full}}=1$ to a full review, $\alpha$ to an assisted verification, and $c_{\mathrm{err}}=10$ to an incorrect automatic diagnosis. A correct automatic diagnosis costs zero. The expected per-wafer cost is
\begin{equation}
\bar{C}(\alpha)
=
\frac{1}{N_{\mathrm{test}}}
\left[
c_{\mathrm{err}}
\!\!\sum_{(x,y)\in\mathcal{A}}\!\!
\mathbf{1}\bigl[\widehat{y}_1(x)\neq y\bigr]
\;+\;
\alpha\,|\mathcal{Q}|
\;+\;
c_{\mathrm{full}}\,|\mathcal{F}|
\right],
\label{eq:supp_cost}
\end{equation}
evaluated at $\alpha\in\{0.1,0.2,0.3,0.5\}$. The confidence-only policy is scored by the same expression with $\mathcal{Q}=\emptyset$, so that every non-automatic wafer contributes $c_{\mathrm{full}}$.

The coefficients are relative rather than absolute: they encode that an erroneous automatic decision is an order of magnitude more costly than a review, and that verifying a two-element set costs a fraction $\alpha$ of an unconstrained review. They are not calibrated to a specific fabrication workflow, and the analysis is used only to compare policies under a common set of penalties. Note also that \eqref{eq:supp_cost} charges $\alpha$ for every assisted wafer regardless of whether the annotated class lies in the returned pair; the reliability of the assisted output is reported separately through \eqref{eq:supp_bucket_top2} rather than folded into the cost.

\section*{Supplementary Material D: A CE-Based Variant That Excludes \texttt{Nonpattern} Pairs}
\label{supp:gate_overlap}

The CE-based approach places \texttt{Nonpattern} in three quarters of its two-class outputs, whereas our framework never does. This section reports a variant of the CE-based approach in which that difference is removed by construction: any returned pair containing \texttt{Nonpattern} is discarded and the wafer escalated instead. Its classifier and its confidence gate are unchanged.

Both approaches then return two-class sets that name two defect patterns, so their outputs become comparable in composition. The variant is not a published method; we include it to characterise what the morphology gate does once the composition difference is set aside.

Table~\ref{tab:gate_overlap} compares what the two approaches admit to assisted diagnosis. Our framework admits $669$ wafers per seed against $414$ for the variant, an increase of $62\%$.

\begin{table}[H]
\centering
\caption{Diagnostic action returned for the same wafer by the two approaches, using ResNet-34 and averaged over four seeds. The CE-based variant excludes any pair containing \texttt{Nonpattern}. Each row is a disjoint group of wafers; counts are seed means rounded to the nearest integer.}
\label{tab:gate_overlap}
\small
\setlength{\tabcolsep}{6pt}
\begin{tabular}{llr}
\hline
Action under CE $+$ confidence & Action under our framework & Wafers \\
(\texttt{Nonpattern} pairs excluded) & & \\
\hline
Assisted diagnosis  & Assisted diagnosis                    & 290 \\
Full review         & Assisted diagnosis                    & 173 \\
Automatic diagnosis & Assisted diagnosis                    & 206 \\
Assisted diagnosis  & Automatic diagnosis or full review    & 124 \\
\hline
\end{tabular}

\vspace{7pt}

\begin{tabular}{lr}
\hline
\multicolumn{2}{l}{\textit{Two-class sets added by the morphology gate (row 2 above)}} \\
\hline
\{\texttt{Edge-Loc}, \texttt{Loc}\}        & 45 \\
\{\texttt{Edge-Loc}, \texttt{Edge-Ring}\}  & 28 \\
\{\texttt{Loc}, \texttt{Scratch}\}         & 28 \\
\{\texttt{Center}, \texttt{Loc}\}          & 21 \\
\{\texttt{Edge-Loc}, \texttt{Scratch}\}    & 11 \\
\{\texttt{Loc}, \texttt{Random}\}          & 10 \\
Sixteen further pairs                      & 29 \\
\hline
Total                                      & 173 \\
\hline
\end{tabular}

\vspace{2pt}
{\footnotesize In the first group the two approaches return the same pair for $273$ of the $290$ wafers. In the fourth group our framework returns a single label for $93\%$ of the wafers and escalates the remaining $7\%$.}
\end{table}

The $173$ wafers of the second group are those the morphology gate admits where the confidence gate escalates. In each of them the two leading classes form a pair that the ambiguity matrix rates as similar, which is the condition the gate tests. Four pairs account for more than two thirds of the set: \{\texttt{Edge-Loc}, \texttt{Loc}\}, \{\texttt{Edge-Loc}, \texttt{Edge-Ring}\}, \{\texttt{Loc}, \texttt{Scratch}\}, and \{\texttt{Center}, \texttt{Loc}\}. These are the same class pairs that motivated the problem in Figure~\ref{fig:taxonomy_ambiguity}. Each of these wafers carries a defect that lies between two named patterns, and the classifier distributes its probability across both of them. The confidence test fails for that reason. The morphology gate applies a different criterion: it evaluates whether the two leading classes are similar in shape, rather than how concentrated the probability is.

The third group is of a different kind. For those $206$ wafers the CE classifier is confident enough to diagnose automatically while the morphology-trained classifier is not, so the difference reflects the two classifiers rather than the two gates. The soft targets of \S B.6 place $0.96$ rather than $1.0$ on the annotated class, which lowers top-1 confidence and moves some wafers below $\tau_{\mathrm{conf}}$. This group is reported for completeness and is not evidence about the gating criterion.

In the other direction, the variant admits $124$ wafers that our framework does not. On $93\%$ of them our classifier is confident enough to return a single label, so no two-class set is required. The remaining $7\%$ are escalated.

\end{document}